\documentclass[final]{cvpr}

\usepackage{times}
\usepackage[utf8]{inputenc} 
\usepackage[T1]{fontenc}    
\usepackage{url}            
\usepackage{booktabs}       
\usepackage{amsfonts}       
\usepackage{nicefrac}       
\usepackage{microtype}      
\usepackage{amsmath}
\usepackage{xcolor}
\usepackage{graphicx}
\usepackage{bigdelim}
\usepackage{floatrow}
\usepackage{subfig}

\usepackage{amsmath, latexsym}
\usepackage{amsfonts}
\usepackage{mathtools}
\usepackage{amsthm}
\usepackage{bm}
\usepackage{xspace}
\usepackage{accents}
\newcommand{\squishlist}{
   \begin{list}{$\bullet$}
    { \setlength{\itemsep}{0pt}      \setlength{\parsep}{3pt}
      \setlength{\topsep}{3pt}       \setlength{\partopsep}{0pt}
      \setlength{\leftmargin}{1.5em} \setlength{\labelwidth}{1em}
      \setlength{\labelsep}{0.5em} } }

\newcommand{\squishlisttwo}{
   \begin{list}{$\bullet$}
    { \setlength{\itemsep}{0pt}    \setlength{\parsep}{0pt}
      \setlength{\topsep}{0pt}     \setlength{\partopsep}{0pt}
      \setlength{\leftmargin}{2em} \setlength{\labelwidth}{1.5em}
      \setlength{\labelsep}{0.5em} } }

\newcommand{\squishend}{
    \end{list}  }

\newfloatcommand{capbtabbox}{table}[][\FBwidth]

\DeclareMathAlphabet{\mathpzc}{OT1}{pzc}{m}{n}

\newcommand{\sideways}{\textit{Sideways}\xspace}
\newcommand{\skipsideways}{\textit{Skip-Sideways}\xspace}

\newcommand{\uniform}{Full-Res\xspace}

\newcommand{\bx}{\bm{x}}

\newcommand{\by}{\bm{y}}

\newcommand{\bh}{\bm{h}}

\newcommand{\bth}{\bm{\theta}}
\newcommand{\timestep}{computation step\xspace}
\newcommand{\timesteps}{computation steps\xspace}

\newcommand{\loss}{\mathcal{L}}
\newcommand{\model}{\mathcal{M}_{\theta}}
\newcommand{\modulefun}{H}

\newcommand{\jacobigt}[2]{\frac{\partial #1}{\partial #2}}

\newcommand{\bxt}[1][]{\ifthenelse{\equal{#1}{}}{\bx^{[t]}}{\bx^{[#1]}}}
\newcommand{\bht}[1][]{\ifthenelse{\equal{#1}{}}{\bm{h}^{[t]}}{\bm{h}^{[#1]}}}
\newcommand{\byt}[1][]{ \ifthenelse{\equal{#1}{}}{\by^{[t]}}{\by^{[#1]}}}
\newcommand{\outt}[1][]{ \ifthenelse{\equal{#1}{}}{\bm{o}^{[t]}}{\bm{o}^{[#1]}}}
\newcommand{\out}[1][]{ \ifthenelse{\equal{#1}{}}{\bm{w}^{[t]}}{\bm{w}^{[#1]}}}
\newcommand{\lt}[1][]{ \ifthenelse{\equal{#1}{}}{l^{[t]}}{l^{[#1]}}}
\newcommand{\bgt}[1][]{ \ifthenelse{\equal{#1}{}}{\bm{\gamma}^{[t]}}{\bm{\gamma}^{[#1]}}}

\newcommand{\pseudograd}{\widetilde{\nabla}}

\newcommand{\ndepth}{D}
\newcommand{\pnumber}{p}

\newcommand{\RR}{\mathbb{R}}

\newsavebox\MBox

\definecolor{darkgreen}{rgb}{0.02, 0.4, 0.12}
\definecolor{dandelion}{rgb}{0.62, 0.52, 0.00}

\usepackage[pagebackref=true,breaklinks=true,colorlinks,bookmarks=false]{hyperref}
\usepackage{xcolor}

\def\confYear{CVPR 2021}

\begin{document}

\title{
Gradient Forward-Propagation for Large-Scale Temporal Video Modelling}

\author{Mateusz Malinowski\\
\small{mateuszm@deepmind.com}\\
DeepMind, U.K.
\and Dimitrios Vytiniotis\\
\small{dvytin@deepmind.com}\\
DeepMind, U.K.
\and Grzegorz \'{S}wirszcz\\
\small{swirszcz@deepmind.com}\\
DeepMind, U.K.
\and Viorica P\u{a}tr\u{a}ucean\\
\small{viorica@deepmind.com}\\
DeepMind, U.K.
\and Jo\~{a}o Carreira\\
\small{joaoluis@deepmind.com}\\
DeepMind, U.K.
}

\maketitle

%
%
\begin{abstract}

How can neural networks be trained on large-volume temporal data efficiently?
To compute the gradients required to update parameters, backpropagation blocks computations until the forward and backward passes are completed. For temporal signals, this introduces high latency and hinders real-time learning. It also creates a coupling between consecutive layers, which limits model parallelism and increases memory consumption. 
In this paper, we build upon \sideways, which avoids blocking by propagating approximate gradients forward in time, and we propose mechanisms for temporal integration of information based on different variants of skip connections. We also show how to decouple computation and delegate individual neural modules to different devices, allowing distributed and parallel training. The proposed \skipsideways achieves low latency training, model parallelism, and, importantly, is capable of extracting temporal features, leading to more stable training and
improved performance on real-world action recognition video datasets such as HMDB51, UCF101, and the large-scale Kinetics-600. Finally, we also show that models trained with \skipsideways generate better future frames than \sideways models, and hence they can better utilize motion cues.
\end{abstract}

\section{Introduction}
Popular deep video models generally rely on spatio-temporal convolutional networks (3D CNNs)~\cite{carreira2017quo,feichtenhofer2019slowfast,stroud2020d3d,tran2015learning,xie2017rethinking} or recurrent networks~\cite{dwibedi2018temporal,hochreiter1997long,li2018videolstm} that are trained with backpropagation (BP) using stochastic gradient descent (SGD)~\cite{bottou2004large, goyal2017accurate, keskar2016large, lecun1988theoretical, robbins1951stochastic, werbos:bptt, werbos1982applications, Werbos:1994:RBO:175610, zhang2004solving}. This is a powerful training paradigm but also an expensive one as it needs to store all the activations in memory to compute Jacobian tensors for the gradient calculations. First, all the activations are computed in the forward mode, from the beginning to the end of the sequence. Next, gradients are computed in the reverse direction, from the end to the beginning. All that severely limits the scale at which we can train temporal models on large-volume sequences such as videos. Therefore, typically these video models are trained on short video clips (about 2.5s at usual frame rate), in offline batch mode.

\sideways~\cite{malinowski2020sideways} is a recent training technique for video models that decouples the computation along the depth of the network and introduces \emph{\timesteps}.  As a new frame is fed into the processing pipeline, each layer independently updates the internal state of the network by computing new activations and gradients. Next, these are passed to the layers above and below in the next \timestep (see Figure~\ref{fig:sideways_skipsideways} left). Moreover, the information cannot be backpropagated to the same units that produced the activations as this happened in the past \timesteps.
One may say that ``everything flows forward in time'', including the backward pass. Due to these properties, \sideways is more biologically plausible than the regular backprop, as it does not block the computation and respects the arrow of time.

Although \sideways, as originally proposed, operates in a temporal forward fashion, it is not a temporal model \emph{per se}, as it has access only to the present frame at each time step. This results in improved memory efficiency similar to single-frame models, making it suitable for real-time applications. However, this comes with the cost of not integrating information temporally, which limits the expressive power of the resulting models.

In this work, we show that it is possible for models to process one frame at a time similar to \sideways, while still being able to extract temporal features. We do this by introducing shortcut (skip) connections in addition to direct connections between the layers of the model. We study the training dynamics of the resulting training procedure, which we call \skipsideways, and show that shortcut connections lead to more stable training and higher accuracy. 

 Regular skip connections~\cite{he2016identity} alter the information flow along the \emph{data} path of the network by allowing activations to `skip' layers, creating data shortcuts. In the proposed \skipsideways, activations and gradients along the shortcut connections are also sent forward in time, effectively creating data paths across time, making it possible to extract temporal features. This change not only extends the modelling space, which subsumes spatio-temporal models, but also gives an interesting perspective on shortcut connections that are typically associated with \emph{vanishing or exploding gradients}~\cite{he2016deep,hochreiter2001gradient}, or \emph{ensembles}~\cite{veit2016residual}.

 To validate the proposed setting, we train a traditional image architecture, \eg, VGG~\cite{simonyan2014very}, on action recognition datasets~\cite{kuehne2011hmdb,soomro2012ucf101}, including a large-scale one --  Kinetics-600~\cite{carreira2017quo} -- by encapsulating the neural modules in \skipsideways units. Since there is no data dependency between units at any one time step, they can process the data in a depth-asynchronous fashion, maximising parallelism. 
 This results in a significant speed-up and reduced latency. The efficiency of the whole network depends only on the worst-case efficiency of an individual \skipsideways unit.
 
 To the best of our knowledge, \skipsideways is the first alternative to backpropagation, more biologically plausible, that is successfully used to train video models at scale.

\section{Related Work}
\label{sec:related_work}
\noindent
\textbf{Pipelined training.}
We are inspired  by recent works on streaming rollouts~\cite{fischer2018streaming,kugele2020efficient}, pipelining during inference~\cite{eccv2018massively} and training~\cite{DBLP:journals/corr/abs-1811-06965, malinowski2020sideways}. In particular, \sideways~\cite{malinowski2020sideways} derives approximate backpropagation rules by taking advantage of the smoothness of the input signal. Even though the authors show  that training is possible even with the mismatch between gradients and activations, their method is unable to learn about scene dynamics during training and inference, and has not yet been demonstrated outside of small-scale settings. \textit{GPipe} ~\cite{DBLP:journals/corr/abs-1811-06965} is another training procedure that does exact backprop by splitting batches into micro-batches, allowing more efficient training. However, it maintains a `bubble' to ensure gradient correctness (Figure 2 in~\cite{DBLP:journals/corr/abs-1811-06965}), hence it cannot use all the hardware resources in parallel, it needs to store activations, still blocks input frames, and  due to the `bubble' the performance is not the same across all frames in the sequence.

\noindent\textbf{Shortcut connections.}  Skip connections have become ubiquitous design elements in modern neural networks~\cite{briggs2000multigrid,campos2017skip,he2016deep,he2016identity,huang2017densely,ripley2007pattern,srivastava2015highway,szeliski2006locally}. These modules were originally introduced to improve training dynamics, by allowing layers to learn residual representations \wrt their preceding layers. This alleviates the modelling burden and results in better propagation of gradients to shallower layers through shortcut connections. Here, the skip connections do not only shortcut between layers along depth, but also along time, by sending activations forward in time~\cite{eccv2018massively,kugele2020efficient,lin2019tsm}. In this way, we can turn frame-based models into spatio-temporal ones without introducing recurrent units.

\noindent\textbf{Alternatives to backpropagation (BP).}
Even though BP is the most commonly used method for training neural networks, multiple works have shown its limitations. BP's biological plausibility is questionable~\cite{betti2019backprop,betti2018backpropagation,kubilius2018cornet,lillicrap2016random}, as the chain rule blocks computations during the forward and backward passes and uses symmetric weights in both passes~\cite{NIPS2016_6441} -- this operating regime is unlikely to exist in biological systems.
Moreover, this also results in high-latency and low throughput in training and deployment of temporal models~\cite{eccv2018massively,malinowski2020sideways}, as it  unnecessarily couples consecutive layers of the neural network architecture, limiting model parallelism and increasing both memory and energy consumption. 

Various decoupling strategies~\cite{balduzzi2014kickback,belilovsky2019decoupled,bengio2007greedy,bengio2015towards,eccv2018massively,choromanska2018beyond,Czarnecki2017UnderstandingSG,huo2018training,lowe2019putting,malinowski2020sideways,NIPS2016_6441,nokland2019training} deal with some shortcomings of BP, but are often applied to visually simpler domains, or static images, or require extra buffers, or investigate solely a biological aspect. 
Here, we are interested in decoupling rules that work well at scale and are applicable to temporally smooth input signals such as videos, that have slowly varying features~\cite{wiskott2002slow}.

\noindent\textbf{Spatio-temporal models.}
Spatio-temporal convolutional neural networks (3D CNNs)~\cite{carreira2017quo,feichtenhofer2019slowfast,stroud2020d3d,tran2015learning,xie2017rethinking} have become  popular recently, mostly due to their performance. However, since they process the whole sequence at once, they are unsuitable for real-time applications and are expensive for representing longer videos, as their memory footprint and training latency grow with the length of the sequence when trained with BP.
Recurrent neural networks (RNN)~\cite{dwibedi2018temporal,hochreiter1997long,li2018videolstm} aggregate information over time and are more suitable for online and causal settings. However, the same memory and latency problems encountered when training 3D CNNs with BP are also present here as RNNs require costly backprop-through-time (BPTT) for training~\cite{robinson1987utility,werbos:bptt,werbos1988generalization}. BPTT scales linearly with the length of the input signal and requires propagation of information forward and backward in time. 
Likewise, causal 3D CNNs~\cite{oord2016wavenet} 
still uses costly BP training that has to be done offline.
In our work, we investigate the setting where: (1) models learn in real-time,  (2) with a direct access only to present data frame, (3) the information (activations, gradients) is sent only forward in time, (4) the training latency is low, 
and (5) we can distribute and parallelize computations.

Recent works~\cite{betti2019backprop,eccv2018massively,hinton2021represent,malinowski2020sideways} model temporal connectivities between modules by connecting consecutive layers at different time steps. Such a structure has implications on learning representation~\cite{betti2019backprop,hinton2021represent} and efficiency~\cite{eccv2018massively,malinowski2020sideways} as it allows depth-parallel computations of activations and pseudo-gradients. Our work belongs to the latter category and we leave learning representation as a  future direction.

\noindent\textbf{Forward propagation in time.} 
Regular training of temporal models with BP or BPTT suffers from the \emph{timing problem}. The common assumption is that the computation of activations and gradients and their propagation are instantaneous~\cite{betti2018backpropagation,kugele2020efficient,malinowski2020sideways} -- this is not physically possible. Real Time Recurrent Learning (RTRL)~\cite{menick2020practical,mujika2018approximating,williams1989learning} and \sideways~\cite{malinowski2020sideways} attempt to mitigate this issue. RTRL computes correct gradients in the forward mode. However, this approach scales poorly for larger input signals as it requires costly high-rank tensor computations. \sideways and our \skipsideways, on the other hand, use approximate gradients in the derivations.

\section{Background}
\label{sec:background}
\begin{figure*}[ht!]
\begin{center}
\begin{tabular}{c@{\ }c@{\ }c}
\hspace{-0.5cm} \includegraphics[width=0.4\linewidth]{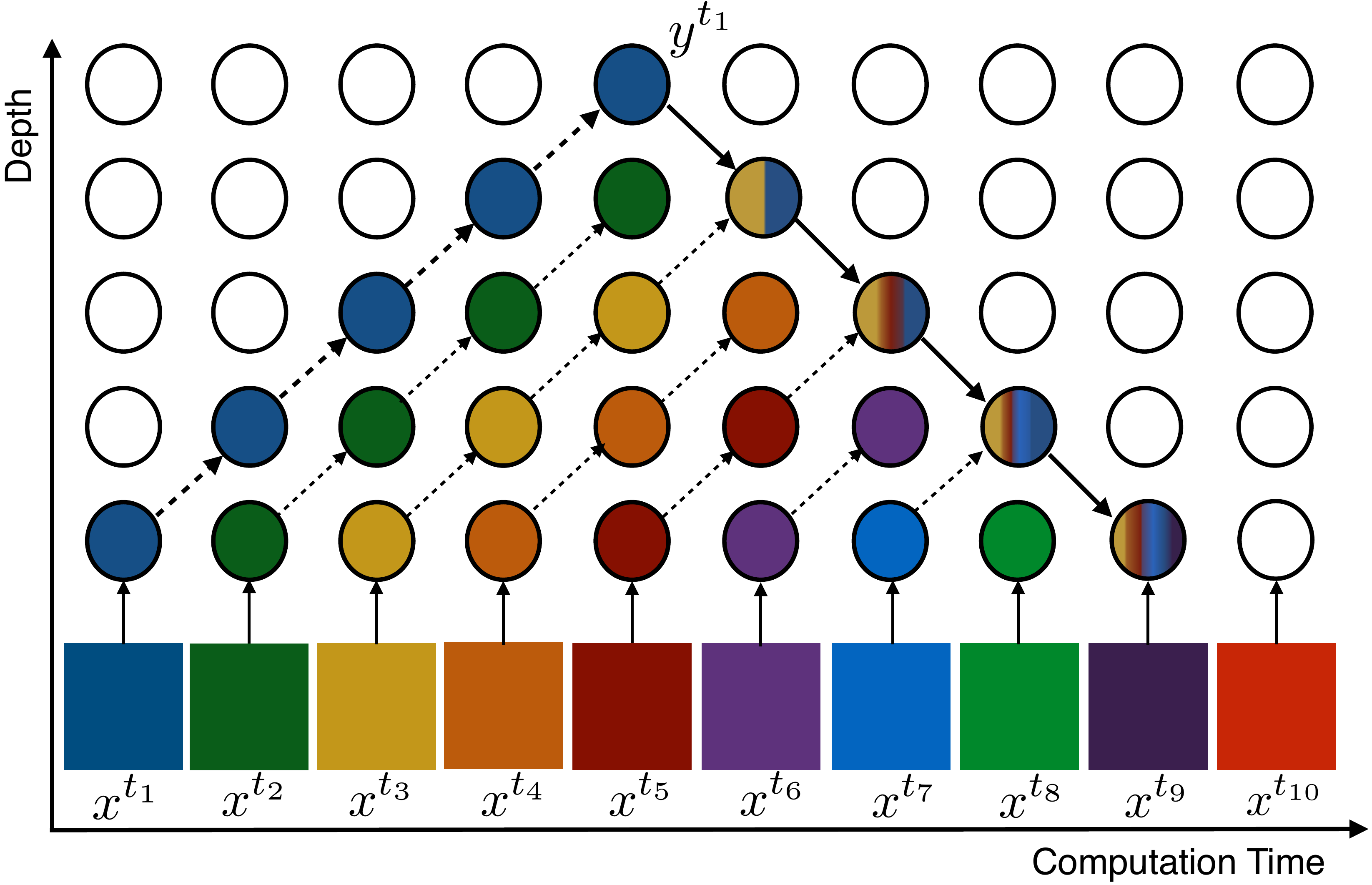} & \hspace{0.5cm}
\includegraphics[width=0.4\linewidth]{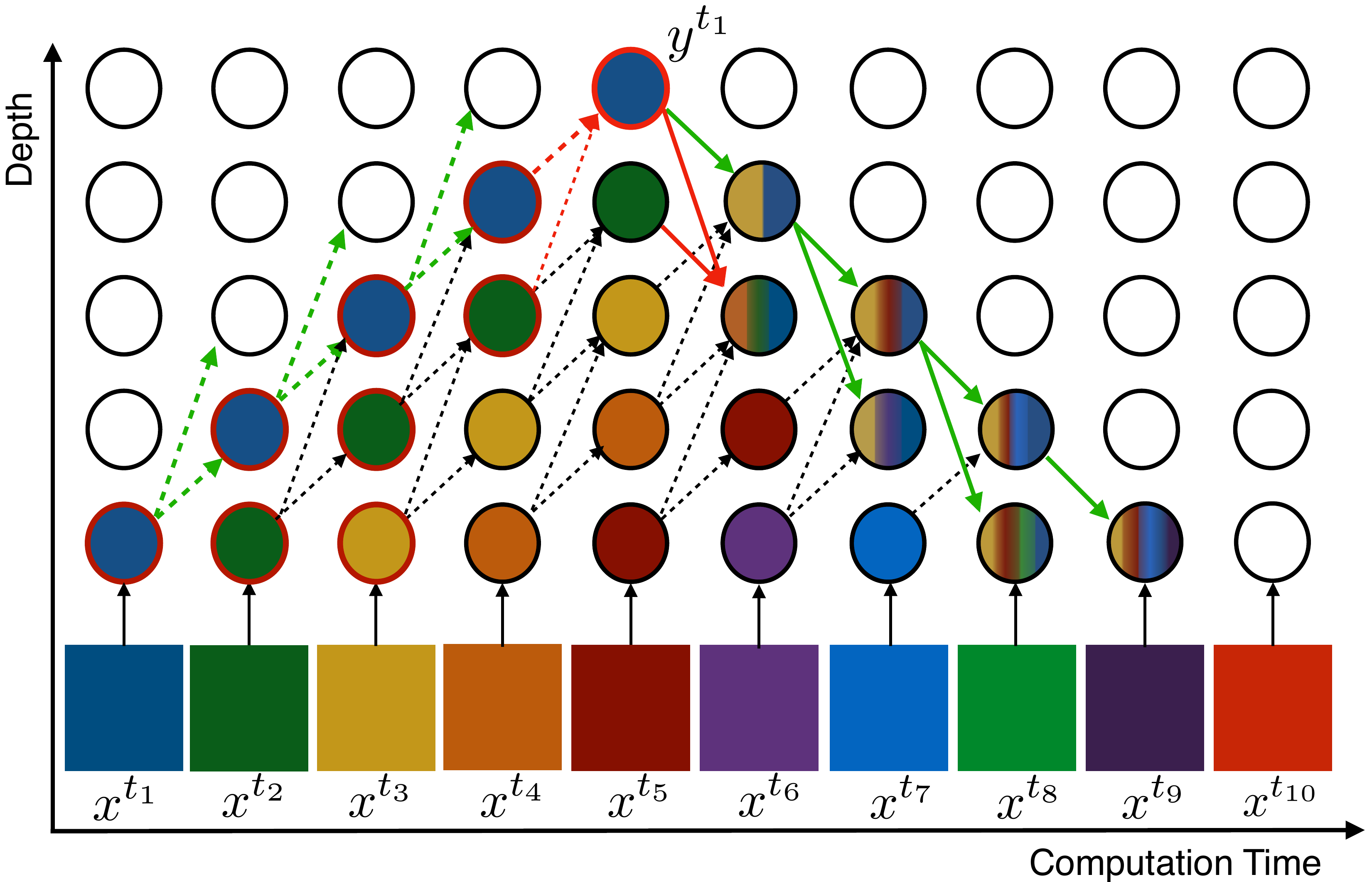}
\end{tabular}
\end{center}
\caption{
\sideways (left) and our \skipsideways (right) unrolled over \timesteps. 
The coloured squares indicate data frames. Circles indicate neural modules inside \sideways or \skipsideways units. We exemplify a fragment that corresponds to the forward pass from $\bxt[t_1]$ (dotted arrows) along with the backward pass (solid arrows). 
In \skipsideways, green arrows illustrate that the information is sent to different units in the next \timestep. Red arrows illustrate how information is integrated from different units in the previous \timestep. Empty circles indicate units outside of the exemplified fragment. Circles with a single colour indicate where the information has originated from. Circles with multiple colours indicate how the information is combined together in the backward pass (we only exemplify a few such units). Individual colours of these multi-coloured circles indicate where the information has originated from. For instance, combining `blue gradient' with `yellow activations' yields `blue-yellow gradient' (6th \timestep and second-last unit). Circles with the red border show the temporal receptive field of the top unit during the forward pass in \skipsideways.
Best viewed in color. 
}
\label{fig:sideways_skipsideways}
\end{figure*}
This section formalizes the learning problem and briefly describes \sideways~\cite{malinowski2020sideways} -- the method upon which we build.

\noindent\textbf{Notation.}
We use the same notation as~\cite{malinowski2020sideways}. We denote input temporal sequences by $\bx = (\bxt)_{t=1}^K, \bxt \in \RR^d$ and outputs by $\by = (\byt)_{t=1}^K, \byt \in \RR^{d_y}$. To compute Jacobians we use the following notation.
Let $\jacobigt{h_l}{h_{l-1}}(\bh_{l-1}, \bth_l) = \left. \frac{\partial \modulefun(h, \bth)}{\partial h} \right|_{h = \bh}$ be the Jacobian tensor of $H(h, \theta)$ with respect to the variable $h$ evaluated at $h=\bh$, $\theta = \bth$. 
Note that, we use bold fonts to distinguish between formal and actual values of the variables. We present complete notation in the supplementary material.

\noindent\textbf{Real-time learning.}
We are mostly interested in efficient, low-latency training, suitable for real-time applications. 
Therefore, we propose the following real-time constraints on our learning problem.
We assume that the model has no direct access to future data, \ie causal setting, and has limited direct access to its past inputs. That is, at time-step $t$ the learning problem is just as a function of $\{(\bxt[t-k], \byt[t-k]), \ldots, (\bxt[t], \byt[t])\}$. In this work, we only investigate the case with $k=0$. 
Moreover, training needs to also scale gracefully to longer sequences, so ideally the memory consumption is independent of the inputs sequence length, and the training does not block computations allowing processing the stream of input data at sufficiently high frame-rate without dropping or buffering input frames.
In our case, we use $25$fps, which is common in video modelling.

\noindent\textbf{Reverse and forward automatic differentiation.} 
BP implements reverse mode for gradient calculation. This allows for efficient gradient calculations by keeping all the tensors low-rank.
More precisely, in the reverse mode, we always combine low-rank gradients with higher-rank Jacobian tensors yielding low-rank outputs. That is, if $\nabla_{h_l}l \in \mathbb{R}^n$ and  $\jacobigt{h_l}{h_{l-1}}(\bh_{l-1}, \bth_l) \in \mathbb{R}^{n^2}$, we have $\nabla_{h_l}l\jacobigt{h_l}{h_{l-1}}(\bh_{l-1}, \bth_l) \in \mathbb{R}^{n}$ as opposed to the forward mode
$\jacobigt{h_l}{h_{l-1}}(\bh_{l-1}, \bth_l)\jacobigt{h_{l+1}}{h_{l}}(\bh_{l}, \bth_{l+1}) \in \mathbb{R}^{n^2}$ implemented by RTRL~\cite{williams1989learning}. The downside of BP is that it has to store in memory the activations over time until the gradients/Jacobians can be computed.

\noindent\textbf{\sideways}
 alters the chain rule of regular BP. It does so by taking into account the passage of time to evaluate individual components of the chain rule at different time steps. The easiest way to think about it is through the composition of two functions. BP assumes the whole process is instantaneous and, in practice, it blocks computations. Therefore, if $F = G \circ H (\bxt[1])$, we have $\jacobigt{f}{x}(\bxt[1]) = \jacobigt{g}{h}(H(\bxt[1])) \cdot \jacobigt{h}{x}(\bxt[1])$. In this way BP combines Jacobians of the \emph{same} input frame. In contrast, \sideways~\cite{malinowski2020sideways} avoids blocking computations by combining Jacobians of \emph{different} input frames and computing \textit{approximate} gradients; we will refer to these approximate gradients as \textit{pseudo-gradients} in the following. For the composition above, \sideways computes $\jacobigt{f}{x}(\bxt[1]) = \jacobigt{g}{h}(H(\bxt[1]))\cdot\jacobigt{h}{x}(\bxt[2])$, so that two inputs can be processed at the same time, and it approximates BP if $\bxt[1] \approx \bxt[2]$. Figure~\ref{fig:sideways_skipsideways} (left) illustrates the idea. Note that since there are no vertical arrows to introduce dependencies, all the layers of the network can be executed in parallel using an appropriate distributed hardware setting. The error introduced by this scheme is difficult to formally quantify. \sideways relies on the fact that the input sequences are smooth and are sampled with a sufficiently high frame-rate, which was empirically shown to hold for real videos and hence validating \sideways. We assume the same here.

\section{\skipsideways}
\label{sec:skipsideways}
\begin{figure}[ht!]
\begin{center}
\begin{tabular}{c@{\ }c@{\ }c}
\hspace{-0.5cm} \includegraphics[width=0.5\linewidth]{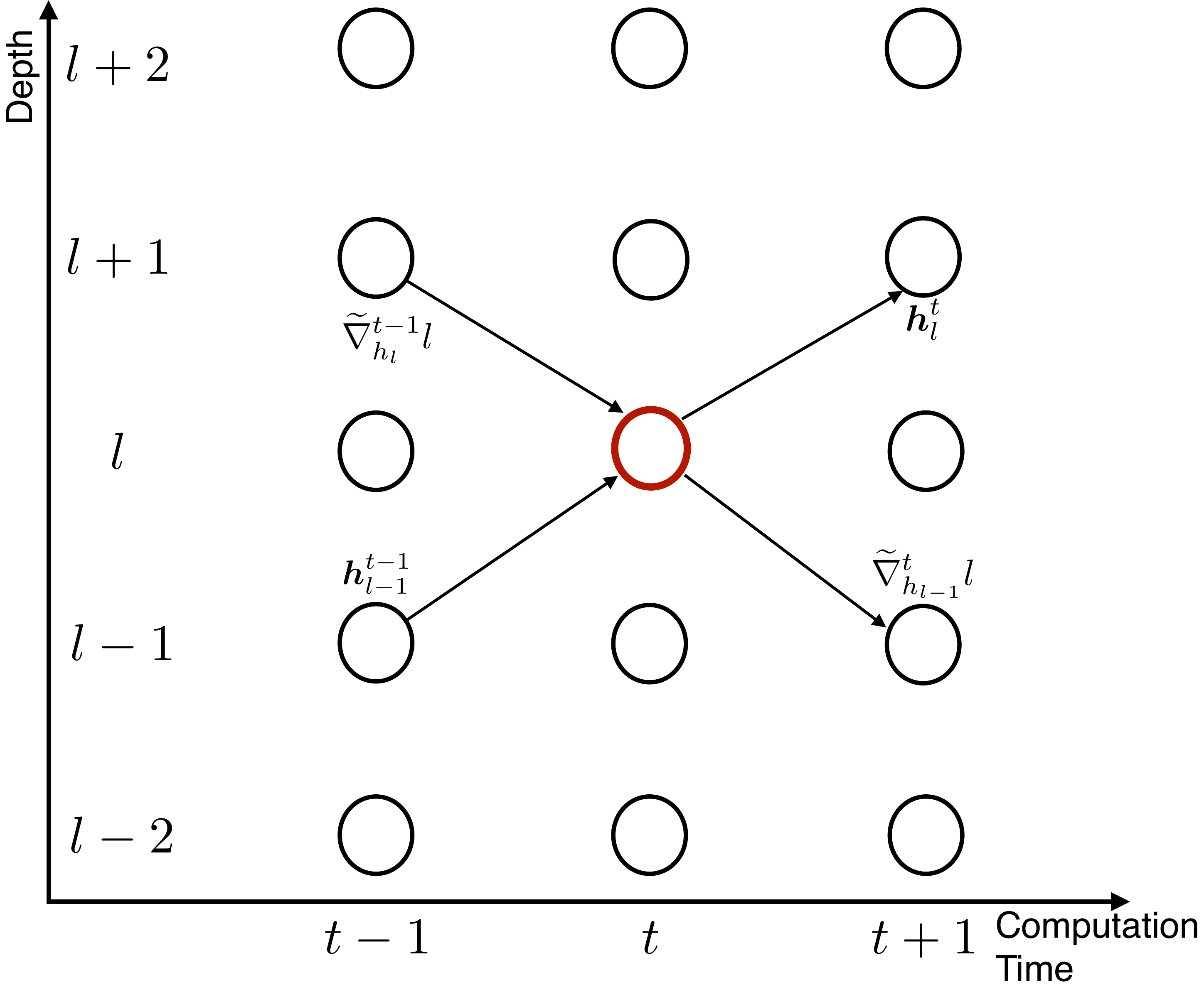} & \hspace{0.5cm}
\includegraphics[width=0.5\linewidth]{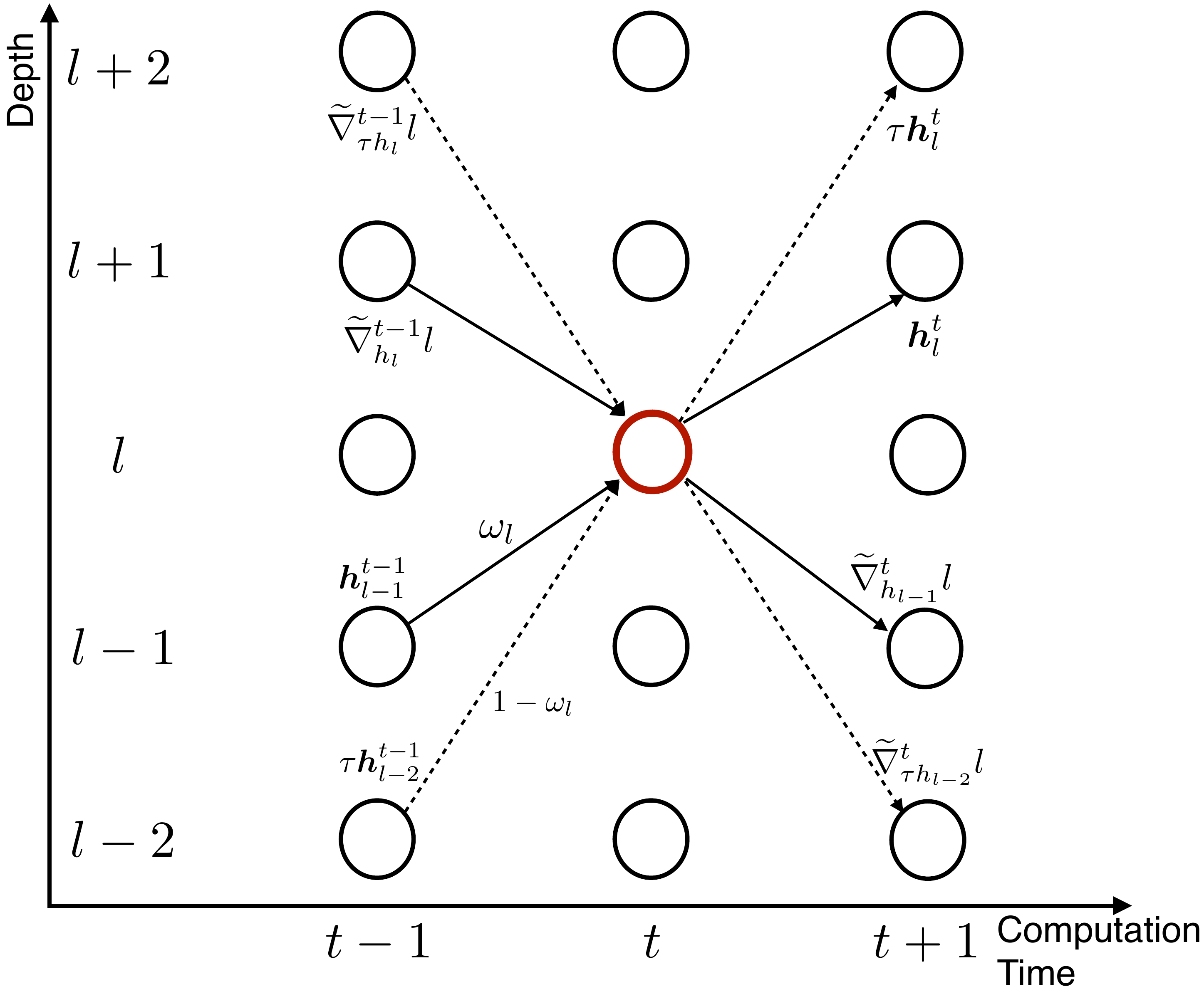}
\end{tabular}
\end{center}
\caption{
Illustration of the data flow of \sideways (left) and \skipsideways (right).
The red circle in the center represents a single unit. It takes activations and pseudo-gradients from the previous step as inputs, processes them, and sends new activations and pseudo-gradients to the next \timestep.
}
\label{fig:skipsideway_closeup}
\end{figure}

Single-frame (2D) CNNs do not have access to motion information other than perhaps through motion blur or semantics. The same holds for \sideways \footnote{Note that, due to \sideways connectivity, the model does combine some information from different frames by using Jacobians originating in different frames within one training iteration (see Figure~\ref{fig:sideways_skipsideways} left). However, this is a weak form of integration and it happens only during the backprop phase.}. This narrow temporal receptive field motivates us to extend \sideways by introducing shortcut connections. Figures~\ref{fig:sideways_skipsideways} and~\ref{fig:skipsideway_closeup} illustrate the difference between \sideways and \skipsideways. We can see that due to the temporal aggregation happening in the forward propagation of both activations and pseudo-gradients through the shortcut connections, individual units have indirect access to more frames. Moreover, their receptive fields increase linearly with the depth of the network, similarly to 3D CNNs with stride $1$~\cite{carreira2017quo,ji20123d,taylor2010convolutional,tran2015learning}. 
We formalise the forward propagation of activations and pseudo-gradients with shortcut connections in the next paragraphs.

\noindent\textbf{Forward-propagation of activations (FA).}
\skipsideways has two types of connections: direct connections, similar to \sideways depicted in Figure~\ref{fig:skipsideway_closeup} with solid arrows, and shortcut connections shown in Figure~\ref{fig:skipsideway_closeup} right, with dotted arrows. In this setting, one unit can receive information from several shallower units, \eg, to pool activations from two units, following formula is used:
\begin{align}
\label{eq:gamma}
\bgt_l = \bht[t-1]_{l-1} \oplus \tau\bht[t-1]_{l-2}
\end{align}
where $\oplus$ is a fusion operator (either addition or concatenation), and $\tau$ denotes the operation to match dimensions between $\bh_{l-1} \in \RR^{d_{l-1}}$ and $\bh_{l-2} \in \RR^{d_{l-2}}$, \eg, max-pool or tile. We can also extend the notation above to $k$ units, \ie, 
\[
\bgt_l = 
\bht[t-1]_{l-1} \oplus \tau\bht[t-1]_{l-2} \oplus \ldots \oplus \tau^{k-1}\bht[t-1]_{l-k}
\]
where $\tau^2 = \tau \circ \tau$ and $\circ$ is composition. For the sake of clarity, we keep our discussion constrained to two blocks only.  Note that \sideways uses a special case when $\bgt_l = \bht[t-1]_{l-1}$.

The choice of fusion operator $\oplus$ in Equation~\ref{eq:gamma} subtly influences the training behaviour. Both addition and concatenation are used in the literature, \eg, addition in ResNets~\cite{he2016identity}, and concatenation in U-NET~\cite{RFB15a}. Addition is a commutative operation which might prevent the network from differentiating between activations coming from direct or shortcut connections. Concatenation, on the other hand, does not suffer from this ambiguity and allows the network to learn appropriate weights to aggregate the information coming from direct and shortcut connections, respectively. Note also that concatenation is well suited to pool over units with different numbers of channels as $\bh_l \oplus \bh_{k} \in \RR^{d_l} \bigoplus \RR^{d_k} \cong \RR^{d_l+d_k}$ and therefore we do not need to perform extra operations to match the channel dimension; this comes however with an increase in the number of model parameters. We report experiments and ablations for both fusion options.

\noindent\textbf{Forward propagation of pseudo-gradients.} 
To account for the aggregation of activations through schortcut connections, we distribute pseudo-gradients using shortcut connections that mirror their counterparts.
Figure~\ref{fig:skipsideway_closeup} shows that a single \sideways unit (red circle in the left figure) takes activations $\bht[t-1]_{l-1}$ and pseudo-gradient $\pseudograd^{t-1}_{l}l$ as inputs, combines them, and outputs new activations $\bht_l$ and pseudo-gradients $\pseudograd^t_{l-1}l$ that are next sent towards the future \timesteps upwards and downwards, respectively. 

In \skipsideways (Figure~\ref{fig:skipsideway_closeup}, right), we integrate activations from direct and shortcut connections, $\bht[t-1]_{l-1}, \tau\bht[t-1]_{l-2}$ together with pseudo-gradients $\pseudograd^{t-1}_{h_l}l$ and $\pseudograd^{t-1}_{\tau h_l}l$.
 Next, the neural unit processes such information and redistributes activations through direct and shortcut connections, $\bht_l$ and $\tau\bht_l$, together with the corresponding pseudo-gradients, $\pseudograd^t_{h_{l-1}}l$ and $\pseudograd^t_{\tau h_{l-2}}l$.

Using total derivative, we can write formally the forward propagation of pseudo-gradients used in \skipsideways:
\begin{eqnarray}
 \pseudograd_{\gamma_l}^t l =& \left(\pseudograd^{t-1}_{h_l} l + \pseudograd^{t-1}_{\tau h_{l}} l \right) \cdot \jacobigt{h_l}{\gamma_l}(\bht[t-1]_{l-1}, \bht[t-1]_{l-2}, \bth_l) \\
 \label{eq:skipsideways_backprop_backpass_1}
   \pseudograd_{h_{l-1}}^t l =& \pseudograd_{\gamma_l}^{t} l \cdot \jacobigt{\gamma_l}{h_{l-1}}(\bht[t-1]_{l-1}, \bht[t-1]_{l-2})\\
  \label{eq:skipsideways_backprop_backpass_2}
   \pseudograd_{\tau h_{l-2}}^t l =& \pseudograd_{\gamma_l}^{t} l \cdot \jacobigt{\gamma_l}{h_{l-2}}(\bht[t-1]_{l-1}, \bht[t-1]_{l-2})\\
  \label{eq:skipsideways_backprop_weight}
  \pseudograd_{\theta_l}^t l =&\left(\pseudograd^{t-1}_{h_l} l + \pseudograd^{t-1}_{\tau h_{l}} l \right) \cdot \jacobigt{h_l}{\theta_l}(\bht[t-1]_{l-1}, \bht[t-1]_{l-2},\bth_{l})
\end{eqnarray}
where rules~\ref{eq:skipsideways_backprop_backpass_1} and~\ref{eq:skipsideways_backprop_backpass_2} `redistribute' the pseudo-gradients during their forward propagation.

Note that we deliberately define shortcut connections 
in the way that the approximation comes only from the temporal misalignment between pseudo-gradients and Jacobians also present in \sideways.  

\section{Experiments}
\noindent\textbf{Tasks and datasets.}
We show the benefits of using the proposed setting \skipsideways over BP and \sideways using two tasks: action recognition and future frame prediction, and on three datasets: HMDB51~\cite{kuehne2011hmdb}, UCF101~\cite{soomro2012ucf101}, and Kinetics-600 (K600)~\cite{carreira2017quo}.
In the action recognition task, a single action, represented as a class, is present throughout an entire video sequence, \ie, $\byt = \byt[t-1]$ for all $t$. 
In the future frame prediction task, the ground truth output $\byt$ changes at each time-step $t$. 
All videos have frame-rate of about 25fps, so that they satisfy  our assumptions regarding smoothness and slowly varying features. HMDB51 and UCF101 are used in \sideways, so for comparison, we use these same datasets in our ablation studies. To showcase our method \skipsideways in large-scale settings, we use Kinetics-600 dataset. We use resolutions $112\times112$ and $224\times224$ and train on $64$ frames for HMDB51, UCF101, and on $32$ frames for K600, with standard input augmentation~\cite{carreira2017quo}.

As we are interested in an alternative to BP that is more scalable and suitable for video modeling, we train all the models from scratch; using features pre-trained with standard BP would defeat the goal of the study. Better results are reported in the literature, but those works use non-causal settings, or use extra features such as flow, extra modalities, different backbones, or various pre-training techniques. These directions are orthogonal to our work, where we investigate alternatives to BP on temporal sequences, in the real-time setting, with no access to future frames. 

\noindent\textbf{Distributed setup.}
For our experiments, we use a distributed hardware with $8$ devices, and we split the network modules among these available devices. As visual backbones in \sideways, \skipsideways, and in baselines trained with standard BP, we use VGG8~\cite{simonyan2014very} for the HMDB51 and UCF101 experiments to have direct comparison with \sideways, and a deeper VGG16 for the Kinetics-600 (K600) experiments. VGG8, having 8 modules, runs as a fully-parallel model when distributed on our hardware, whereas for VGG16, we assign pairs of consecutive modules to every device. The computation for the modules on the same device happens in sequence using regular BP, and the \skipsideways direct and shortcuts connections are applied between devices. We provide more details about the experimental setting in the supplementary material.

\noindent\textbf{Baselines.}
We have implemented multiple baselines for our experiments. First, we re-implemented causal 3D VGG16 based on~\cite{oord2016wavenet}. This baseline is close to our \skipsideways as it does not see future frames and it is capable of extracting temporal features. However, it is trained with regular BP.

As results for \sideways on large-scale datasets like Kinetics-600 are not reported in the literature, we have re-implemented \sideways. We have obtained slightly better results than those reported in the original paper~\cite{malinowski2020sideways}. We include the results from our implementation throughout this section.

For comparison with existing works in terms of hardware efficiency, we also implemented a video model using GPipe-style training~\cite{DBLP:journals/corr/abs-1811-06965}. We provide the details of the implementation in the supplementary material.

For the reconstruction task, we use a simple 16-layer deep convolutional model, without pooling or striding, hence maintaining full-resolution throughout the depth of the model; we denote this model as \uniform. We depart from the traditional encoder-decoder design for two main reasons. First, we hypothesize that preserving higher spatial resolution is important for a dense prediction task like reconstruction. Second, and more interesting in the context of our work, even though typically such a design decision would lead to prohibitively expensive (memory- or speed-wise) training, we can readily afford to train such models by distributing \skipsideways units to different devices, using the distributed hardware efficiently. 
We include further details of the architecture in the supplementary material.

\subsection{Results}

\begin{table}[t]
    \centering
    \begin{tabular}{llll}
        \toprule
           & HMDB51 & UCF101 & K600 \\
        \midrule
        \sideways~\cite{malinowski2020sideways}   & 27.3 & 62.5 & 55.7 \\
        \textbf{\skipsideways}   & \textbf{31.4} & \textbf{64.5} & \textbf{61.8} \\
        \bottomrule
    \end{tabular}%
    \caption{
    \skipsideways vs \sideways~\cite{malinowski2020sideways} on HMDB51, UCF101, and Kinetics-600 (K600); resolution $112\times112$. We use VGG8 for HMDB51 and UCF101, and VGG16 for K600. We report accuracy in $\%$. \skipsideways uses concatenation as aggregation operator.
    }
    \label{tab:sideways_vs_skip}
\end{table}

\begin{table}[t]
    \centering
    \begin{tabular}{llll}
        \toprule
        Ablation  & Fusion & HMDB51 & UCF101\\
        \midrule
        \sideways~\cite{malinowski2020sideways}   & - & 27.3 & 62.5 \\
         \skipsideways  & add & 29.6 & 62.8 \\
        \textbf{\skipsideways}  & \textbf{concat} & \textbf{31.4} & \textbf{64.5} \\
         FA-\skipsideways   & concat & 29.7 & 61.9 \\
        \bottomrule
    \end{tabular}%
    \caption{
    Ablations on HMDB51, UCF101; resolution $112\times112$. We use VGG8 for HMDB51 and UCF101. We use addition (add) or concatenation (concat) operator to integrate activations.
    We report accuracy in $\%$.
    }
    \label{tab:ablations}
\end{table}

\begin{table}[tb]
     \begin{tabular}{l|ll|ll}
                    \toprule
                    Ablation & \multicolumn{2}{c|}{HMDB51} & \multicolumn{2}{c}{UCF101} \\
                    & reg. & wide & reg. & wide \\
                    \midrule
                    \sideways~\cite{malinowski2020sideways} & 27.3 & 27.8 & 62.5 & 64.6 \tabularnewline
                    \skipsideways  & 31.4 & 31.5 & 64.5 & 66.0 \tabularnewline
                    \bottomrule
                \end{tabular}%
            
                \caption{
                Regular vs. wide models trained with \sideways or \skipsideways. Wide \sideways has the same number of parameters as \skipsideways. 
                Accuracy of the latter is superior (27.8 vs 31.4). 
                We report accuracy in $\%$.
                \label{tab:ablations_other}
                }
\end{table}

\begin{table}[]
\begin{tabular}{l|c}
                    \toprule
                    Training  & 256 frames \\
                    \midrule
                    \sideways + VGG16 & 40  \\
                    \skipsideways + VGG16 & 40  \\
                    BP + 3D VGG16 & - \\
                    BP + 3D VGG16 (Remat) & 1 \\
                    \bottomrule
                \end{tabular}%
                \caption{
                 The largest batch-size that can be fit into a single GPU under the same setting, for different training schemes. Fitting a batch-size 1 for 3D VGG16 results in out-of-memory (OOM) error. (Remat) uses rematerialization~\cite{kumar2019efficient}.
                \label{tab:batch_size}
                }
\end{table}

\noindent\textbf{\skipsideways vs \sideways.}
Table~\ref{tab:sideways_vs_skip} shows our main results comparing our proposed \skipsideways to \sideways~\cite{malinowski2020sideways}. \skipsideways significantly improves over \sideways, about $4$, $2$, and $6$ per cent points (pp) respectively, showing the benefit of the temporal integration of information.

\noindent
\textbf{Ablations.} Our ablation studies justify different design choices used in \skipsideways and compare the results with \sideways in Table~\ref{tab:ablations}. 

As the first analysis, we investigate the effect of using shortcut connections only in the forward pass to aggregate activations, without distributing the pseudo-gradients as detailed in Section \ref{sec:skipsideways}. We report the results in Table~\ref{tab:ablations}, 4th row, FA-\skipsideways. Although FA-\skipsideways improves the results slightly over \sideways, re-distributing the pseudo-gradients as proposed in our full \skipsideways brings considerable additional improvement.

The choice of fusion operator is important, with a clear preference towards concatenation, \ie, \skipsideways with concatenation outperforms the \sideways baseline by about $4$ and $2$ per cent points (pp), respectively, and on K600 by $6$ pp (see Table~\ref{tab:sideways_vs_skip}).

To rule out that the increased performance of \skipsideways with concatenation is due to the larger number of parameters rather than the temporal integration of information, we run experiments with wider VGG8, so that this wider VGG8 and \skipsideways with concatenation have similar number of parameters. We report these results in Table~\ref{tab:ablations_other}. For HMDB51, we can see larger improvement by $3.6$pp by adding shortcut connections than merely by adding more parameters. The results are less clear for UCF101 where models benefit from more parameters.
Nonetheless, in both cases, training with \skipsideways significantly improves the results over \sideways.

\noindent\textbf{Stability.}
Similar to the original \sideways paper~\cite{malinowski2020sideways}, we study the training stability with respect to the input frame-rate, learning rate, and the number of video cuts (`montage'). We have found that \skipsideways training is slightly more stable than \sideways. Interestingly, lower input frame-rate affects less the \skipsideways performance compared to \sideways, which collapses when the frame rate drops. This robustness shows the benefit of faster data paths obtained through shortcut connections. More details on this analysis are included in the supplementary material (`Stability'), where we also show that the method converges for lower learning rates under reasonable smoothness assumptions.

\noindent\textbf{Memory consumption.} Our training scheme significantly reduces memory consumption compared to the standard BP training. This is because \skipsideways is a temporal training scheme that forward-propagates all the information (activations and pseudo-gradients), without the need to store activations throughout the sequence length like in BPTT. To illustrate this behaviour, we have trained 3D VGG16 (with BP), \sideways, and \skipsideways, and we measured the maximal possible batch-size on a single Quadro P5000 (16GB) in all three settings, for the same sequence length. Table~\ref{tab:batch_size} shows that \sideways and \skipsideways training allows for much larger batches, whereas BP training has resulted in out-of-memory error. This result opens the door to dealing with long videos, beyond the commonly used 2-3 second clips, or to methods that need very large batches such as contrastive methods~\cite{chen2020simple}, or alternative and more expensive architectural designs such as \uniform. Note that \skipsideways has a memory usage similar to \sideways even though, advantageously, the former has a temporal receptive field growing linearly with the number of \skipsideways units, whereas the \sideways's temporal receptive field is one.

\noindent\textbf{Speedup.}
Besides the important memory savings, another important practical advantage of our proposed training mechanism is the increased speedup obtained due to model parallelism. Table~\ref{tab:speed} shows our results where we distribute each \skipsideways unit on a different device (with $8$ devices in total). 
We compare \skipsideways with the regular BP (running on a single device\footnote{Note that, because of the dependencies between layers, even if we distribute BP training on different devices, the results would be slower due to communication overhead.}) and its distributed version, GPipe~\cite{DBLP:journals/corr/abs-1811-06965}, running on 8 devices. 
The number of steps per second reported in Table~\ref{tab:speed} represent the number of whole training iterations per second, including: fetching data\footnote{Our asynchronous data fetching has a negligible cost.}, forward and backward passes, and weight updates. As we can see, \skipsideways is almost three times faster than the regular BP training and about two times faster than GPipe. It is also the only one that satisfies our real-time requirements (Section~\ref{sec:background}) as it can process frames at 25fps, it is causal, and is independent of the sequence length, being able to scale to potentially infinite sequences. Moreover, we also compare our model parallelism with batch parellelism (the last two rows in Table~\ref{tab:speed}), where we train the network with total batch-size 8 distributed among eight devices. 
Note that these parallelism schemes are complementary and  \skipsideways can also be used with batch-parallelism. In the supplementary material we provide more results.

\noindent\textbf{\skipsideways vs BP.}
In Table~\ref{tab:best}, we compare VGG16 trained with \skipsideways to other models trained with BP. Since \skipsideways is a causal model, for fair comparison, we have implemented a causal equivalent of 3D VGG16 and trained it with BP. First, we can see that there is little difference between I3D~\cite{carreira2019short,carreira2017quo}, an established model for action recognition, and a 3D CNN with the VGG16 backbone~\cite{simonyan2014very} that we denote as 3D VGG16 (see 1st and 2nd rows). Hence using 3D VGG16 for action recognition is a sensible choice. Next, we can observe that there is a significant decrease in accuracy when transforming the 3D VGG16 into a causal one (1st and 3rd rows). 
Our \skipsideways (last row) outperforms the causal 3D VGG16 showing that it is a valid and competitive training mechanism, while being real-time. We have also found that using another visual backbone (InceptionNet) gives similar results to VGG16.

Note that the state-of-the-art results on Kinetics-600 are about $83\%$~\cite{feichtenhofer2020x3d,qiu2019learning}. As mentioned above, these results are generally obtained using various mechanisms that are orthogonal to our work. Here, we use standard visual backbones and focus on alternative to BP training that satisfies real-time requirements and is more scalable.

\begin{figure*}[t]
\begin{center}
\begin{tabular}{c@{\ }c@{\ }c@{\ }c@{\ }c@{\ }c@{\ }c}
\rotatebox{90}{\quad\quad\quad Input} & \includegraphics[width=0.17\linewidth]{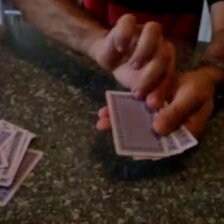} & 
\includegraphics[width=0.17\linewidth]{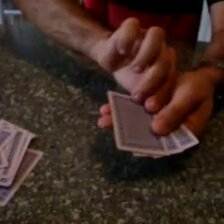} &
\includegraphics[width=0.17\linewidth]{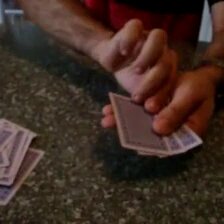} &
\includegraphics[width=0.17\linewidth]{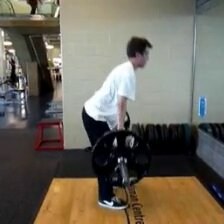} & 
\includegraphics[width=0.17\linewidth]{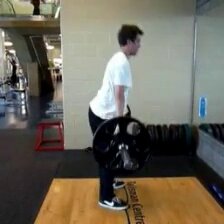} &
\includegraphics[width=0.17\linewidth]{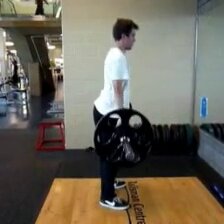}
\\
\rotatebox{90}{\quad Ground-Truth} & \includegraphics[width=0.17\linewidth]{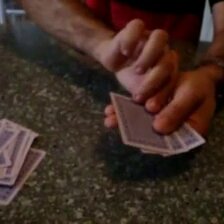} & 
\includegraphics[width=0.17\linewidth]{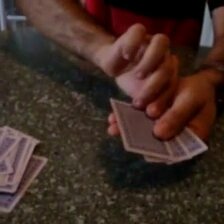} &
\includegraphics[width=0.17\linewidth]{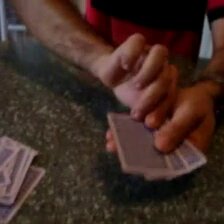} &
\includegraphics[width=0.17\linewidth]{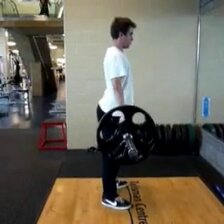} & 
\includegraphics[width=0.17\linewidth]{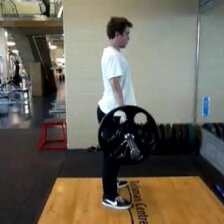} &
\includegraphics[width=0.17\linewidth]{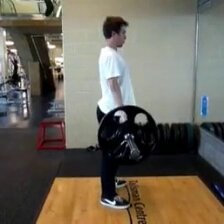} 
\\
\rotatebox{90}{\quad\quad \sideways} &\includegraphics[width=0.17\linewidth]{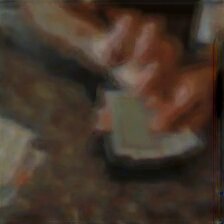} & 
\includegraphics[width=0.17\linewidth]{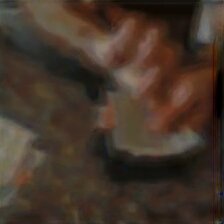} &
\includegraphics[width=0.17\linewidth]{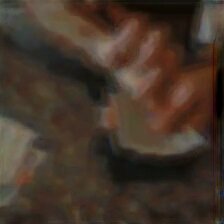} &
\includegraphics[width=0.17\linewidth]{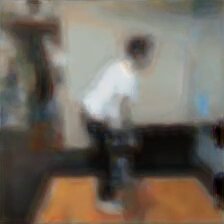} & 
\includegraphics[width=0.17\linewidth]{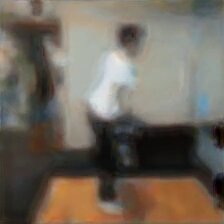} &
\includegraphics[width=0.17\linewidth]{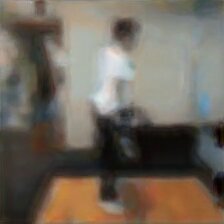} 
\\
\rotatebox{90}{\skipsideways (Ours)} &\includegraphics[width=0.17\linewidth]{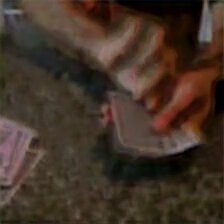} & 
\includegraphics[width=0.17\linewidth]{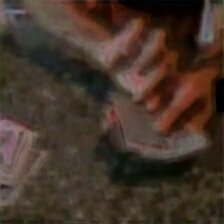} &
\includegraphics[width=0.17\linewidth]{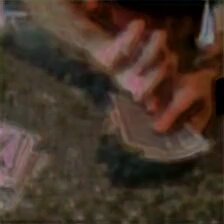} &
\includegraphics[width=0.17\linewidth]{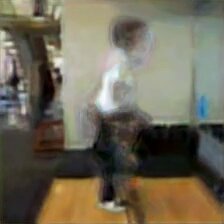} & 
\includegraphics[width=0.17\linewidth]{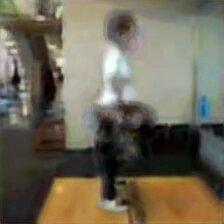} &
\includegraphics[width=0.17\linewidth]{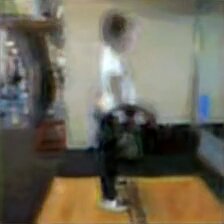}
\end{tabular}
\end{center}
\caption{Qualitative results for two video sequences from the validation fold of Kinetics-600, where the models need to generate the frame at step $t+8$ into future given frames up to $t$. From top to bottom: input frames, ground-truth output frames, predictions of \uniform  trained with \sideways, predictions of \uniform trained with \skipsideways.
We can observe that the predictions of the model trained with \sideways are blurry and semantically similar to inputs, indicating that the model lags behind. On the contrary, the predictions of the model trained with \skipsideways are sharper and semantically closer to the ground-truth output frames. We use resolution $224\times 224$.
}
\label{fig:future_forecasting}
\end{figure*}

\noindent\textbf{Future frame prediction.}
Future frame prediction is our second task.
Figure~\ref{fig:future_forecasting} shows results when training models with \sideways, \skipsideways. As expected, motion is a powerful cue to predict future video frames, and thereof our \skipsideways training significantly outperforms \sideways training in terms of L2-loss  (Table~\ref{tab:future_forecasting}). We can observe that \sideways produces blurry video frames, as it is unable to extract motion features to reduce uncertainty over where the pixels will be moving in future frames. On the contrary, the predictions obtained with \skipsideways are much sharper.

\begin{table}[t]
    \centering
    \begin{tabular}{lll}
        \toprule
          & $\frac{\text{steps}}{\text{sec}}$  & Devices \\
        \midrule
        \skipsideways + VGG16  & $2.3$  & 1 \\
        \skipsideways + VGG16  & $3.6$  & 2 \\
        \skipsideways + VGG16  & \textbf{7.2}  & 8 \\
        GPipe~\cite{DBLP:journals/corr/abs-1811-06965} + VGG16  & 4.2  & 8 \\
        BP + 3D VGG16 & $1.9$ & 1 \\
        BP + 3D VGG16 (Remat) & $1.5$ & 1 \\
        BP + Causal 3D VGG16  & $2.4$ &  1 \\
        BP + I3D~\cite{carreira2017quo}  & $2.9$ &  1 \\
        \midrule
        BP + 3D VGG16 (Batch-paral.) & 5.5 & 8 \\
        BP + I3D (Batch-paral.) & 6.5 & 8 \\
        \bottomrule
    \end{tabular}%
    \caption{
        Speed comparisons of various training strategies and architectures in number of steps per second. 
        Column \textit{devices} indicates the number of parallel devices used for distributed training. All models use K600, $32$ frames, resolution $224\times224$, and batch-size 8. 
        (Remat) denotes rematerialization \cite{kumar2019efficient}.
        The last two rows are exceptions and use batch-size 1 per device. We use space-to-depth transformation.
    }
    \label{tab:speed}
\end{table}

\begin{table}[b]
    \centering
    \begin{tabular}{llll}
        \toprule
         Kinetics-600 & Backbone  & Causal & Accuracy \\
        \midrule
        BP & 3D VGG16  & N & $72.3$ \\
        BP & I3D~\cite{carreira2019short,carreira2017quo} & N & $71.9$ \\
        BP & Causal 3D VGG16 & Y & $64.2$ \\
        \skipsideways & VGG16  & Y & $67.3$ \\
        \bottomrule
    \end{tabular}%
    \caption{
        Results on Kinetics-600. We use resolution $224\times224$ and total batch-size $256$ in our experiments.
    }
    \label{tab:best}
\end{table}

\begin{table}[b]
    \centering
    \begin{tabular}{llll}
        \toprule
          Kinetics-600 & L2-Error \\
        \midrule
        \sideways~\cite{malinowski2020sideways}   &  0.107 \\
        \textbf{\skipsideways}   & \textbf{0.073} \\
        BP & - \\
        \bottomrule
    \end{tabular}%
    \caption{
    Results of \uniform  on Kinetics-600, trained with batch size 64, $112\times112$, to predict future frames at time-step $t+8$, given the frames up to time-step $t$.
    We report pixel error as L2-norm. Training with BP in this configuration results in out-of-memory (OOM) error.
    }
    \label{tab:future_forecasting}
\end{table}

\section{Conclusion}
Alternatives to regular training based on backprop face many challenges as most of the existing frameworks and hardware systems are developed to optimise training based on backprop. 

Yet, even with all the existing optimisations, regular backprop has limitations as it does not scale up with the sequence length. It keeps large buffers of activations or gradients, and couples consecutive layers of the network, which blocks model parallelism along the depth of the model. These result in high latency systems, which may be unsuitable for real-time vision. In this work, we show that there are possible alternatives to backprop that reduce coupling between layers, decrease latency, allow for depth-parallelism, and can integrate information temporally, to obtain accuracy comparable to standard 3D CNNs. Our work is the first to showcase an alternative to backprop at scale and we hope that we will inspire further research into hardware architectures and software implementations that go beyond the standard backprop paradigm. As future work, we will investigate different architecture designs, 
mechanisms to augment the proposed \skipsideways with long-term memory capabilities~\cite{hochreiter1997long}, and investigate similar mechanisms for learning better representations.

\section{Appendix}

In this supplementary material we provide additional details about the experimental setup used in the main paper, results about the stability of the proposed \skipsideways training, and additional qualitative results obtained in the reconstruction task.

\subsection{Notation}
We denote input temporal sequences by $\bx = (\bxt)_{t=1}^K, \bxt \in \RR^d$ and outputs by $\by = (\byt)_{t=1}^K, \byt \in \RR^{d_y}$. We mainly focus on video clips, where each frame has dimension $d = h \times w \times 3$. In action recognition tasks, we have $\byt = \byt[1]$. We use  single-frame 2D CNNs that map inputs to logits $\model : \RR^d \rightarrow \RR^{d_y}$. These networks are compositions of layers, written as
\[
\model(\bxt) = \modulefun_D(\cdot, \theta_\ndepth) \circ \modulefun_{\ndepth-1}(\cdot, \theta_{\ndepth-1}) \circ \ldots \circ \modulefun_1(\bxt, \theta_1)
\]
where each layer $\modulefun_l(\cdot, \cdot)$ is a function $\modulefun_l : \RR^{d_{l-1}} \times \RR^{\pnumber_{l}} \rightarrow \RR^{d_{l}}$, $\circ$ is a composition, \ie, $G \circ F(\bxt) = G(F(\bxt))$, and $\theta=\left(\theta_l\right)_l, \theta_l \in \RR^{\pnumber_l}$ are trainable parameters. We also use $\bht_l = \modulefun_{l}(\cdot, \theta_{l}) \circ \ldots \circ \modulefun_1(\bxt, \theta_1)$ to denote activations at layer $l$.

We use $\loss : \RR^{d_y} \times \RR^{d_y} \rightarrow \RR$ defined as
$\loss(\bh_D, \by) = \sum_{t=1}^K l(\bht_D, \byt) = \sum_t \lt$, where $\bht_D$ are logits, \ie, $\bht_D = \model(\bxt)$, and $l$ is a loss at time $t$, \eg, a cross-entropy loss $l(\bh, \by) = -\sum_i p(\by_i)\log q(\bh_i)$. We update model parameters with the temporally averaged gradients, \ie, $\theta := \theta - \alpha\sum_t \nabla_{\theta}\lt$, where $\alpha$ is a learning rate. 

We  use the following notation for Jacobian tensors.
Let $\jacobigt{H(h, \theta)}{h}(\bh, \bth) = \left. \frac{\partial \modulefun(h, \bth)}{\partial h} \right|_{h = \bh}$ be the Jacobian tensor of $H(h, \theta)$ with respect to the variable $h$ evaluated at $h=\bh$, $\theta = \bth$. 
Note that, we use bold fonts to distinguish between formal and actual values of the variables. We  also use a shorthand notation $\jacobigt{h_l}{h_{l-1}}(\bh_{l-1}, \bth_l) = \jacobigt{H_l(h_{l-1}, \theta_l)}{h_{l-1}}(\bh_{l-1}, \bth_l)$.

\subsection{Experimental Setup}
\noindent\textbf{Datasets.}
We use three datasets:
\begin{itemize}
    \item HMDB51 has 6,770 video clips representing 51 actions~\cite{kuehne2011hmdb}, sampled at 30fps. We use the same train and test splits as~\cite{jing2018self,malinowski2020sideways,simonyan2014two}. 
    \item UCF101~\cite{soomro2012ucf101} has 13,320 videos clips and 101 human actions, sampled at 25fps, with clips having 7.21sec on average. 
    \item Kinetics600 (K600) is a large-scale dataset for action recognition~\cite{carreira2017quo} with around 490,000 videos. It has 600 human actions, and each action has at least 600 video clips per action. The clip duration is 10s, with frame-rate 25fps. 
\end{itemize}

\noindent\textbf{Tasks.} Here we give additional details for the reconstruction task used in the main paper.
At each time-step $t$, the network predicts the frame at time-step $t+8$, \ie, we want to learn a mapping $\mathcal{M}_{\theta}(\bxt; \theta)\rightarrow \byt = \bxt[t+8]$ for all $t$. We choose a displacement of 8 frames into the future as this corresponds to the number of \skipsideways or \sideways units that we use. That is, the output layer has to make a prediction for the frame that the first layer is about to see. Hence the models have no access to the observation they must predict. To successfully solve this task, it may be helpful for the models to learn to capture the underlying motion in the scene so they can re-distribute pixels accordingly.  We minimise the following objective: $\frac{1}{T}\sum_{t=1}^T || \mathcal{M}_{\theta}(\bxt) - \bxt[t+8]||^2_{l_2}$. We compare \skipsideways against \sideways.
The results of \skipsideways are significantly better than the equivalent model trained with \sideways (see additional qualitative results included in Figure~\ref{fig:future_forecasting_1}) and it shows the benefit of integrating temporal information through direct and shortcut connections.

\noindent\textbf{Distributed setting.}
We experiment on K600 using two multi-host setups, where we use either $8$ or $16$ hosts. Each host has $8$ TPUs (Tensor Processing Units) amounting to $64$ TPUs in the first setup, and $128$ in the second setup. We use the second setup only for large batch-size experiments shown in Table 6, in the main paper. Unless we write otherwise, we use batch-size $16$ per host. Hence, the total batch-size is $128$ in the first setup and $256$ in the second setup.

For VGG8, we put each layer of the network into a single \skipsideways unit and stick all units to different devices.
For VGG16, we group two consecutive layers into a single \skipsideways unit and also stick all units to different devices.
Note that VGG's design is convenient for the experiments as it has no shortcut connections and has roughly the same number of layers as the maximal number of devices (one per device for VGG8 and two for VGG16) that we use to distribute \skipsideways units. In the distributed setup, we use JAX~\cite{jax2018github} and Haiku~\cite{haiku2020github}.

\begin{figure*}[t]
\begin{center}
\begin{tabular}{c@{\ }c@{\ }c}
\hspace{-0.6cm} \includegraphics[width=0.56\linewidth]{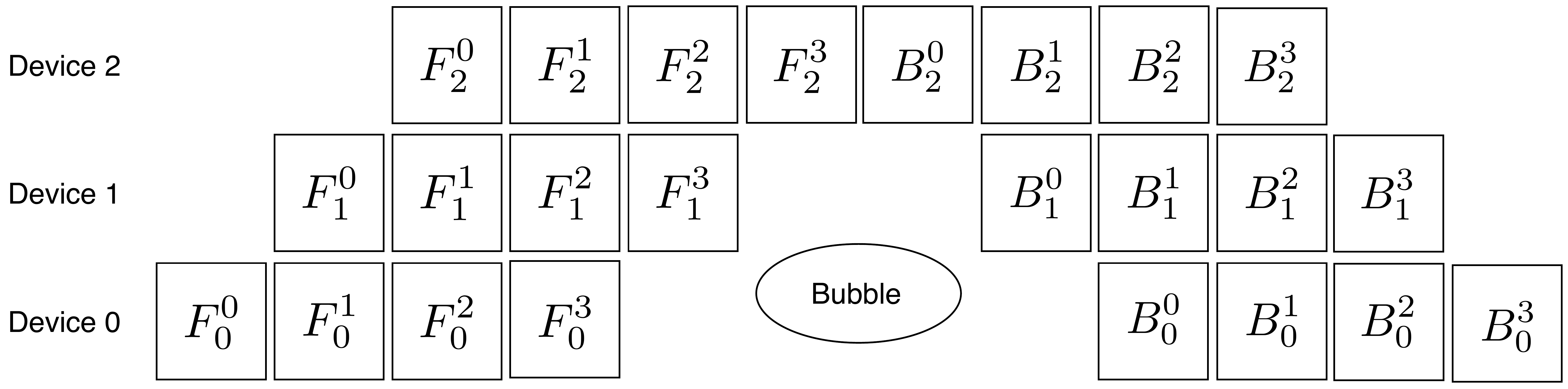} &
\hspace{0.cm} \includegraphics[width=0.5\linewidth]{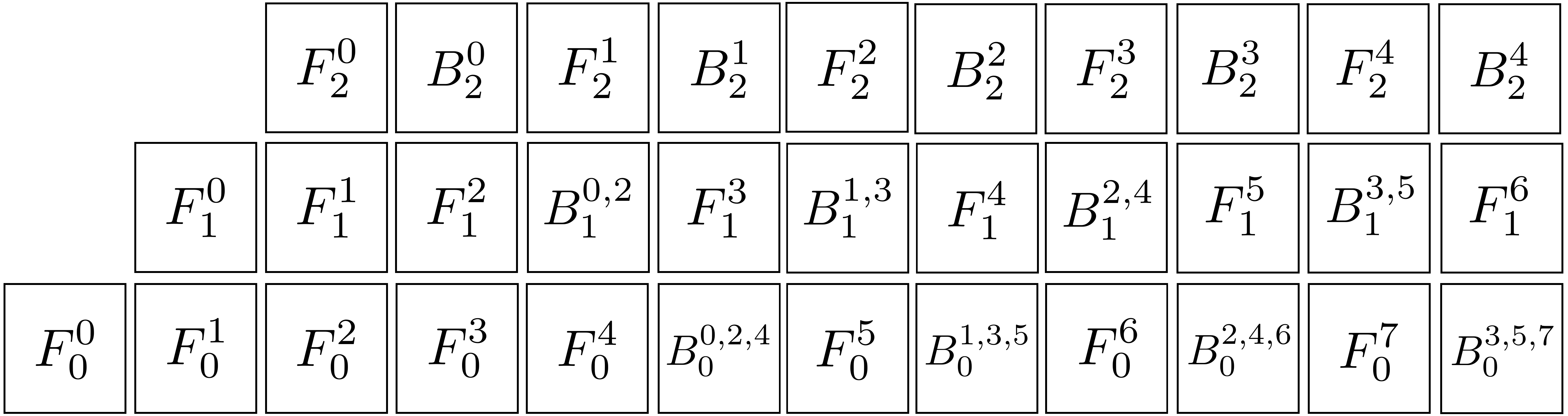}
\end{tabular}
\end{center}
\caption{
We show our adaptation of GPipe~\cite{DBLP:journals/corr/abs-1811-06965} to video modelling (left) and compare it with \sideways~\cite{malinowski2020sideways} and our \skipsideways (right).
We only show the device placement during the forward pass ($F$) and backward pass ($B$). Because of that, we ignore shortcut connections in \skipsideways in this illustration. Here, we consider three distributed devices. $F_k^t$ denotes that activations computed in the forward pass from the $t$-th input frame are placed on the $k$-th device. We use similar notation for gradients during the backward pass, \ie, $B_k^l$. For \sideways and \skipsideways, we use $B_k^{l_1, l_2}$ to denote that the gradient was obtained from the incoming gradient computed based on the $l_1$-th input frame and Jacobian based on the $l_2$-th input frame. We extend this notation to $B_k^{l_1, l_2, l_3}$ that denotes the gradient $B_{k+1}^{l_1, l_2}$ is combined with Jacobian computed based on the $l_3$-th input frame on the $k$-th device. For the sake of clarity, we only show four input frames for GPipe. However, we can see that in the same number of twelve \timesteps, \skipsideways can process more input frames and does not induce the `bubble'. The vertical axis denote different devices and horizontal axis computation time.
}
\label{fig:gpipe_sideways_devices}
\end{figure*}

\noindent\textbf{GPipe.}
We have adapted GPipe~\cite{DBLP:journals/corr/abs-1811-06965}, another realisation of a distributed training by pipelining. 
 Here, we use the same breakdown of our models to different pipeline stages, \eg, two consecutive VGG16 layers are allocated to a single pipeline stage and hence to a single device. Since the original GPipe has been designed to work with non-temporal data, we have extended it to work with videos by creating batches of frames as micro-batches, using the GPipe terminology.
As GPipe implements correct BP, it also has two phases, forward and backward, with a parameters update that follows that on each device separately. Figure~\ref{fig:gpipe_sideways_devices} illustrates comparison between GPipe and \skipsideways training in terms of device allocation.

In summary, in GPipe, (i) every micro-batch induces a forward and a backward computation on every device, (ii) activations on every device have to be kept alive until the backwards pass, hence increasing memory pressure, (iii) unless the number of micro-batches is sufficiently large there exists a non-negligible pipeline `bubble' effect, and (iv) latency, and hence the overall performance, varies as the result of the moving `bubble'. By contrast, (i) we only pay the cost of the backward computation (to simultaneously compute activations
and gradients), (ii) each device does not have to hold previous activations as we exploit temporal correlation, (iii) we induce no bubble, (iv) latency is the same for each frame. 

\subsection{Models}
\begin{figure*}[t]
\begin{center}
\begin{tabular}{c@{\ }c}
\hspace{-0.6cm} \includegraphics[width=0.56\linewidth]{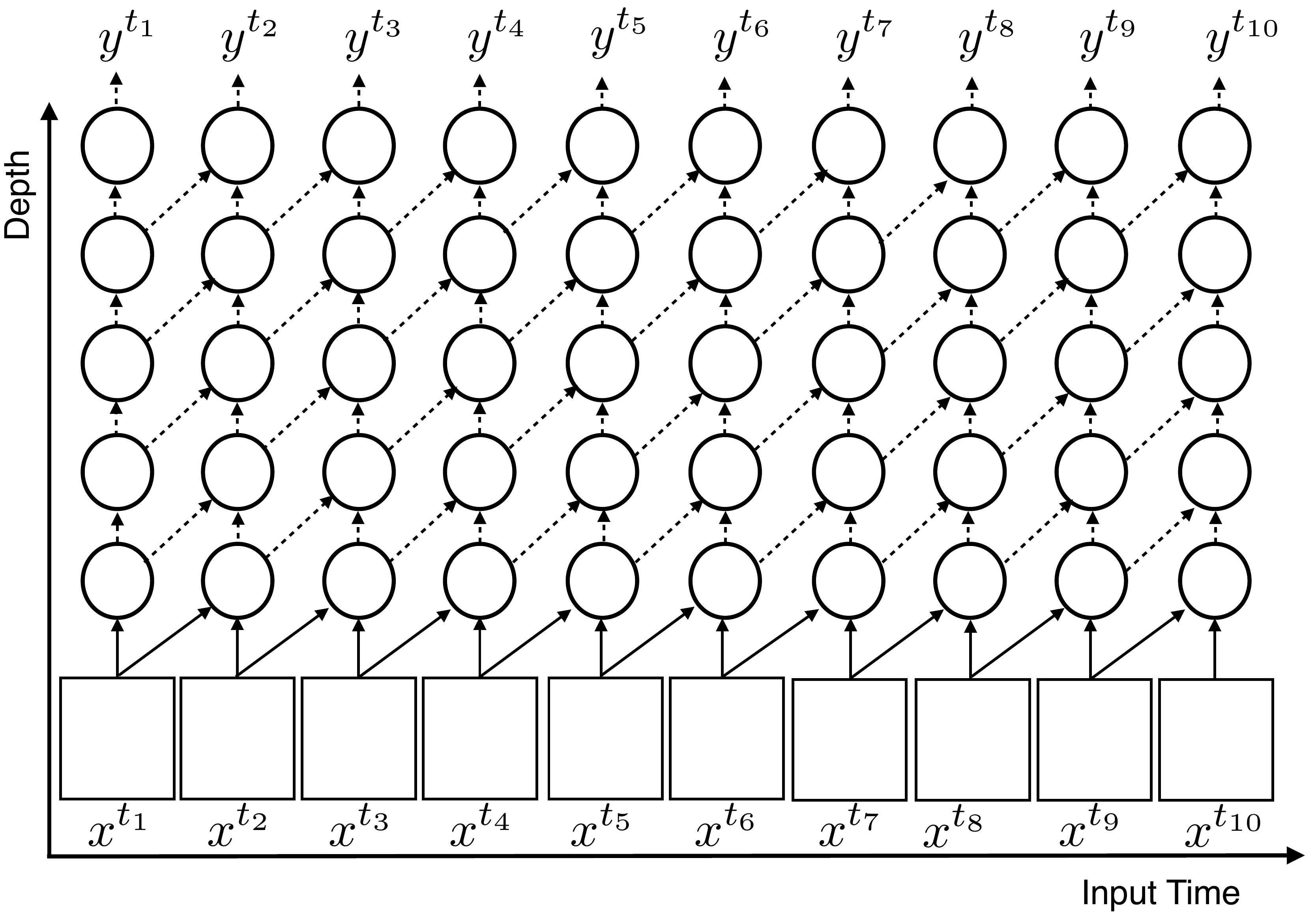} &
\hspace{0.cm} \includegraphics[width=0.5\linewidth]{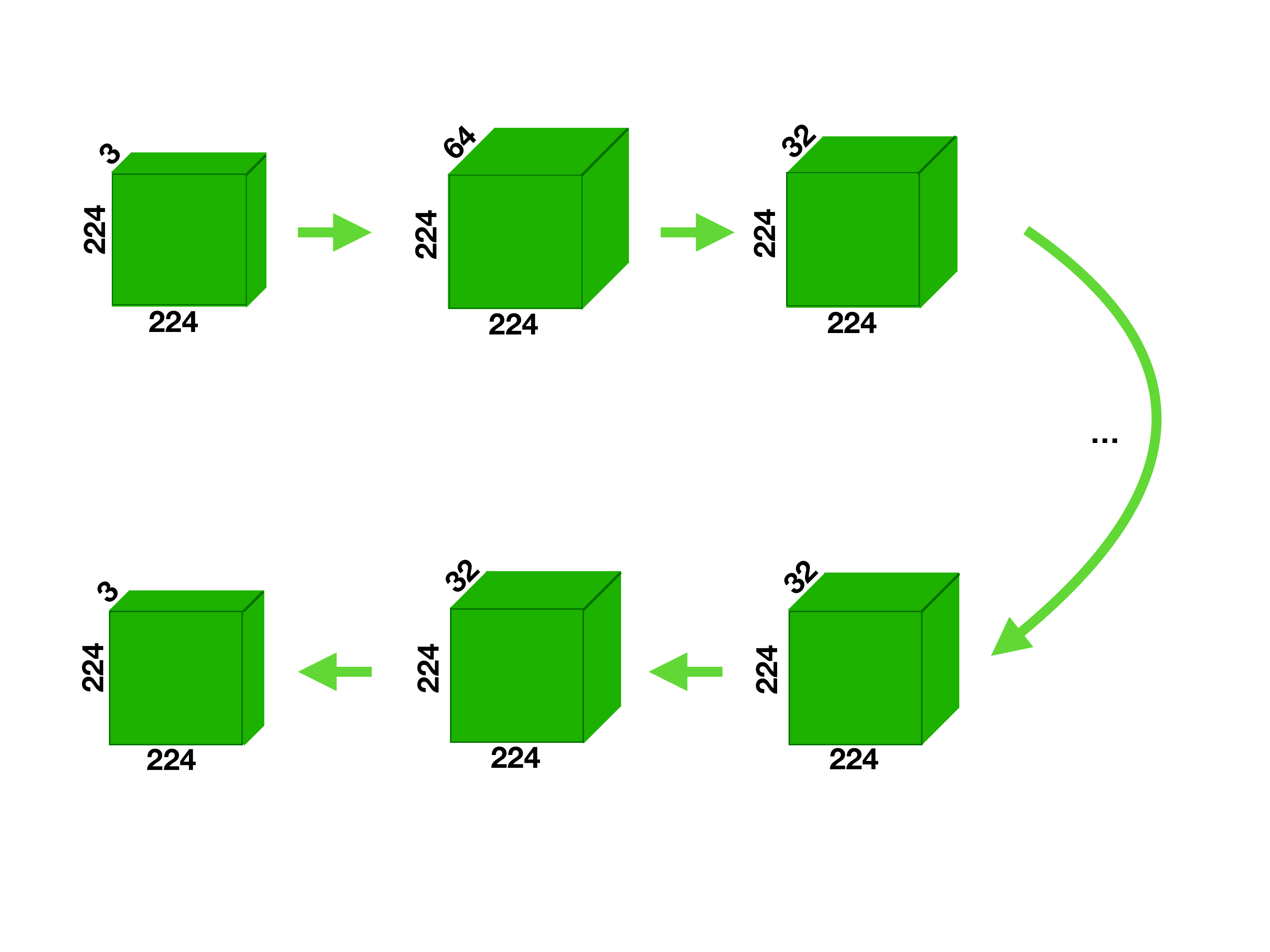} \\
\small a) Causal 3D VGG. & b) \uniform.
\end{tabular}
\end{center}
\caption{
Our two baseline architectures. Causal 3D VGG and \uniform. Note that, contrary to the Figure 1 in the main paper, here we only show the connectivity of the architecture, and not the flow of pseudo-gradients nor activations. Moreover, the time in the figure refers to input data frames -- not the computation time.
}
\label{fig:causal_and_uniform}
\end{figure*}

\noindent\textbf{3D VGG and Causal 3D VGG.} We `inflate'~\cite{carreira2017quo} 2D VGG to 3D VGG and train it from scratch, following the I3D model. That is, we extend spatial VGG16~\cite{simonyan2014very} kernels to include temporal direction. We use $3\times3\times3$ kernel shapes for convolutions and max-pooling. Since \skipsideways training is real-time and hence causal, for a fair comparison, we have also implemented Causal 3D VGG. We have observed a slightly better performance with kernels $3\times3\times3$ over $2\times3\times3$ (from $63.7$ to $64.2$). For simiplicity, we illustrate the later variant in Figure~\ref{fig:causal_and_uniform}a). We train all the models with ADAM~\cite{adam} and a cosine learning rate schedule.

\noindent\textbf{\uniform network.}
As a proof-of-concept, we have implemented a network that maintains the input resolution through all the layers. We have designed the architecture to be (i) conceptually simple, (ii) not reduce spatial dimensions, and (iii) be better at utilising hardware resources in a distributed setting. Even though it better utilises the distributed hardware resources, its training is expensive with BP and using a single device. The network consists of a stack of convolutional kernels with $3\times3$ filters, and pooling layers with the same kernel shape ($3\times$3). In all cases, we use striding one, so that there is no reduction in spatial dimensions. We use $64$ channels in the first two consecutive blocks, and next $32$ channels for rest of the network except from the last layer that maps back the latent space into $3$ channels, which we interpret as RGB channels. We use Rectified Linear Units (ReLU) after each convolution, and hard hyperbolic tangent~\cite{nwankpa2018activation} in the last layer. Each two consecutive convolutional layers are followed by a single pooling layer. We use eight such blocks that we assign to different distributed devices within the \skipsideways units. We do not use a pooling layer after the last convolutional layer. We use batch-normalization just before ReLU. Figure~\ref{fig:causal_and_uniform}b) illustrates the architecture showing all the blocks.

\subsection{Speed and Memory}
We show results on speed improvements and memory savings.

\noindent\textbf{Space-to-depth.}
In all our experiments, we use space-to-depth transformation, which makes computations more TPU-friendly. More precisely, we first divide each frame into 2-by-2 non-overlapping patches and then we align them along the channel dimension. This transformation decreases each spatial dimension by two, and increases the channel dimension by four. Equivalently, it also increases the spatial receptive field of convolutional neural networks two times per spatial dimension and results in fewer but more costly convolutions (see Figure 4 in~\cite{wang2018learning})\footnote{\url{https://www.tensorflow.org/api_docs/python/tf/nn/space_to_depth}}. Overall, however, space-to-depth transformation often speeds up training without affecting the accuracy of the model, but this is hardware and model specific. We also experiment with spacetime-to-depth transformation where we repeat the same procedure also along the time dimension. It further speeds up the computations as shown in Table~\ref{tab:speed_large}. For our experiments with \sideways and \skipsideways, we keep a more pure setup with only space-to-depth transformation. This is because we want to explicitly model the temporal transitions through the \sideways or \skipsideways temporal connections.

\noindent\textbf{Rematerialization.}
Rematerialization~\cite{kumar2019efficient} is a method that saves memory at the cost of extra computation. That is, the technique recomputes some intermediate activations instead of storing them in memory. Specifically, we use `haiku.remat' where we rematerialize each vgg-block consisting of two or three 3D CNNs.

\begin{table}[t]
    \centering
    \begin{tabular}{lll}
        \toprule
          & $\frac{\text{steps}}{\text{sec}}$  & Devices \\
        \midrule
        \skipsideways + VGG16  & $2.3$  & 1 \\
        \skipsideways + VGG16  & $3.6$  & 2 \\
        \skipsideways + VGG16  & \textbf{7.2}  & 8 \\
        GPipe~\cite{DBLP:journals/corr/abs-1811-06965} + VGG16  & 4.2  & 8 \\
        BP + 3D VGG16 & $1.9$ & 1 \\
        BP + 3D VGG16 (Remat) & $1.5$ & 1 \\
        BP + Causal 3D VGG16  & $2.4$ &  1 \\
        BP + I3D~\cite{carreira2017quo}  & $2.9$ &  1 \\
        \midrule
        BP + 3D VGG16 (Batch) & 5.5 & 8 \\
        BP + I3D (Batch) & 6.5 & 8 \\
        BP + I3D (Batch)$^\ast$ & 7.9 & 8 \\
        \bottomrule
    \end{tabular}%
    \caption{
        Speed comparisons of various training strategies and architectures in number of steps per second. 
        Column \textit{devices} indicates the number of parallel devices used for distributed training. All models use K600, $32$ frames, resolution $224\times224$, and batch-size 8.
        (Remat) denotes rematerialization \cite{kumar2019efficient}.
        The last two rows are the exceptions and use batch-size 1 per device.
        We use space-to-depth transformation.
        $^\ast$ denotes space-time-to-depth transformation.
    }
    \label{tab:speed_large}
\end{table}

\subsection{Theoretical Savings}
\sideways and \skipsideways do not propagate gradients back in time, only forward in time. This means that, in contrast to the traditional backpropagation, both mechanisms do not need to store activations over a long sequence. Therefore the memory cost is the same as the cost of doing backpropagation of per-frame models and is \emph{independent of} the number of frames. Hence, as long as the memory savings are concerned, both mechanisms should scale up indefinitely with respect to the duration of the video. Moreover, in a distributed setting, if we do not count the costs of inter-device communication, the overall speed should be bounded only by the speed of the slowest device. That is, if the network is a composition of $D$ modules $\model(\bxt) = \modulefun_D(\cdot, \theta_\ndepth) \circ \modulefun_{\ndepth-1}(\cdot, \theta_{\ndepth-1}) \circ \ldots \circ \modulefun_1(\bxt, \theta_1)$ and each module $\modulefun_j$ sits on the $j$-th device, the run-time $\mathcal{T}$ of $\model$ is $\mathcal{T}(\model) = \max_j \mathcal{T}(\modulefun_j)$. Notice that with regular backprop the same runtime is the sum of runtimes of individual modules.

\subsection{Stability}
\begin{figure*}[hbt]
\begin{center}
\begin{tabular}{c@{\ }c@{\ }c}
\hspace{-0.5cm} \includegraphics[width=0.3\linewidth]{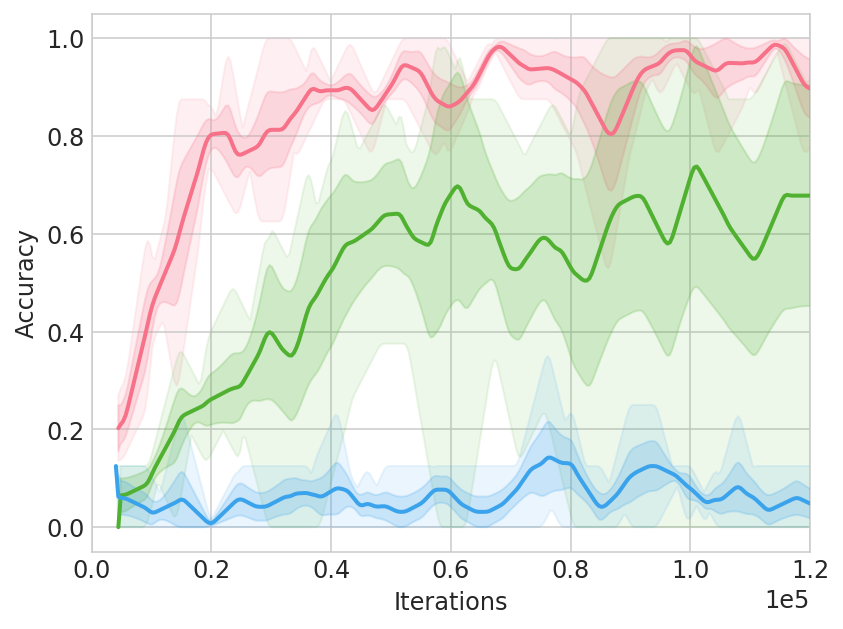} & \hspace{0.5cm}
\includegraphics[width=0.3\linewidth]{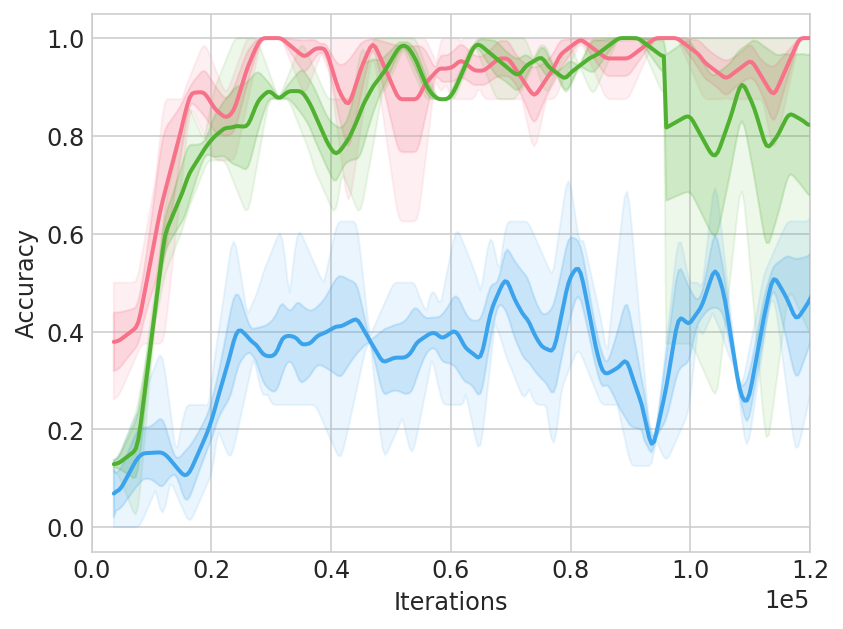}
\end{tabular}
\end{center}
\caption{
Robustness on the training fold with respect to the choice of the initial learning rate for \sideways (left) and our \skipsideways (right).
Red, green and blue denote the initial learning rate as $0.01, 0.1, 1.0$. Solid curves depict the mean and shaded areas show variance across various input frame-rates.
}
\label{fig:robustness_lr}
\end{figure*}

\begin{figure*}[ht]
\begin{center}
\begin{tabular}{c@{\ }c@{\ }c}
\hspace{-0.5cm} \includegraphics[width=0.3\linewidth]{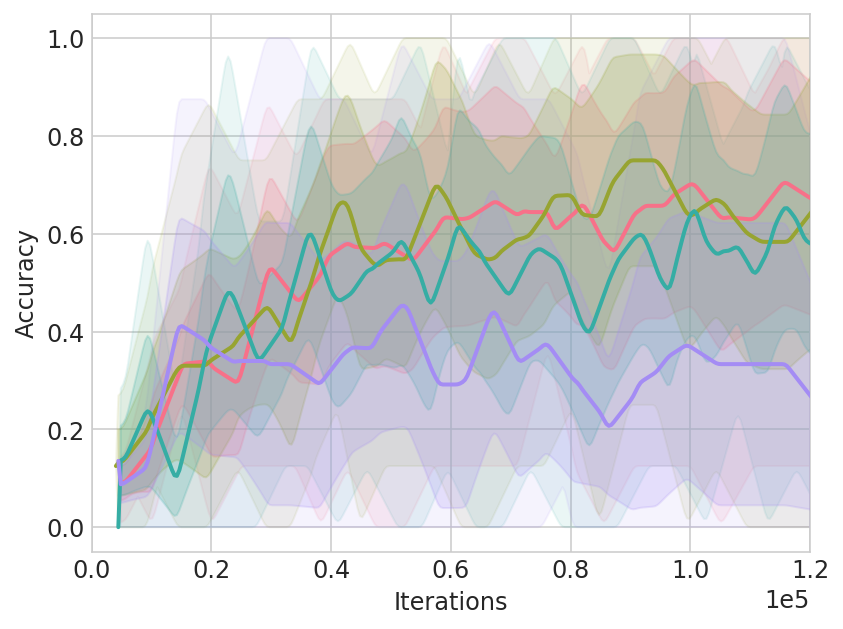} & \hspace{0.5cm}
\includegraphics[width=0.3\linewidth]{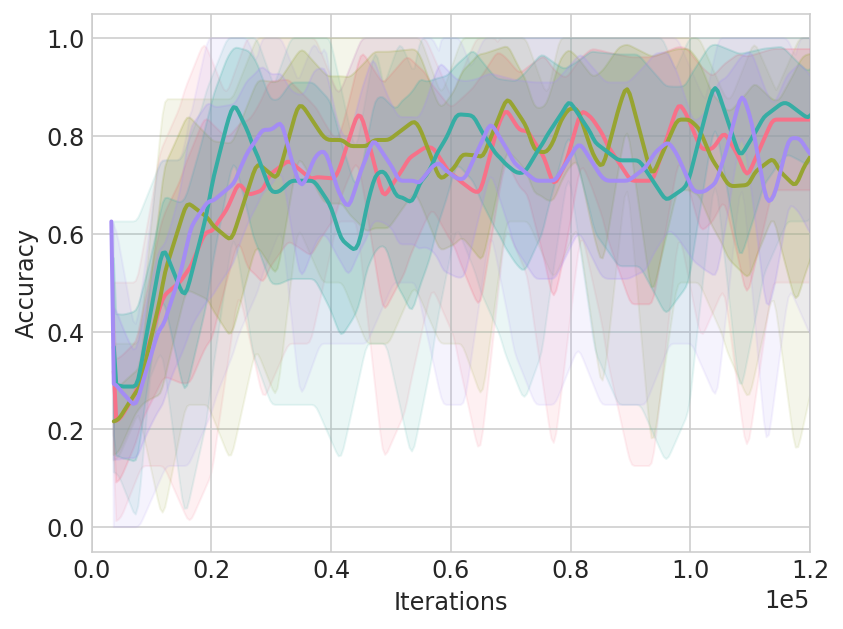}
\end{tabular}
\end{center}
\caption{
Robustness on the training fold with respect to the choice of the input frame-rate for \sideways (left) and our \skipsideways (right).
Red, olive, green and violet denote the input frame-rate as $30, 15, 7, 4$ respectively. Solid curves depict the mean and shaded areas show variance across various input initial learning rates.
}
\label{fig:robustness_frame_rate}
\end{figure*}

\begin{figure*}[ht]
\begin{center}
\begin{tabular}{c@{\ }c@{\ }c}
\hspace{-0.5cm} \includegraphics[width=0.3\linewidth]{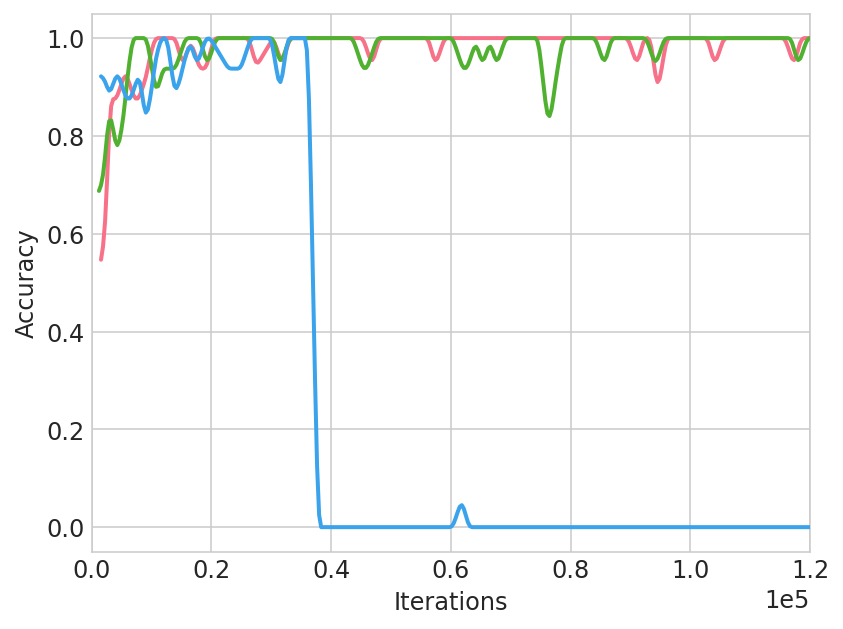} & \hspace{0.5cm}
\includegraphics[width=0.3\linewidth]{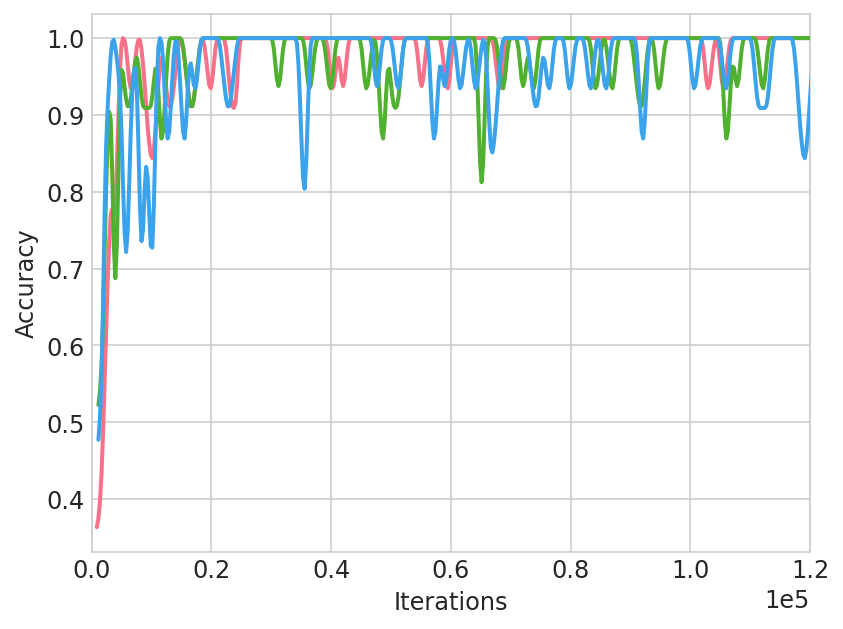}
\end{tabular}
\end{center}
\caption{
Robustness of \skipsideways on the montage experiments where we concatenate two (left) or three (right) video clips into a larger video.
Red, green and blue denote the initial learning rate as $0.0001, 0.001, 0.01$. We can see that in general more video cuts hurt the performance. Sometimes the learning collapses for larger learning rates.
}
\label{fig:skip_sideways_montage_concat}
\end{figure*}

Our method assumes smoothness of the input signal. In this section, we investigate two settings that break that assumption and see how those impact our training.
\newline\noindent\textbf{Frame-rate.}
We run experiments on HMDB51 to investigate the effect of hyperparameters (learning rate and input frame rate) on training stability for the proposed \skipsideways compared to the original \sideways~\cite{malinowski2020sideways}. Intuitively, the higher the learning rate or the lower the frame-rate, the more we depart from our initial assumptions about the smoothness of the input or intermediate activations across time steps, hence the less stable the training should be. For frame-rate, we consider the values $30$fps, $15$fps, $7$fps, $4$fps. For learning rates, we use the values $1.0, 0.1, 0.01$.

Figure~\ref{fig:robustness_lr} shows the behaviour when using different initial learning rates and Figure~\ref{fig:robustness_frame_rate} shows the behaviour when varying the input frame-rate.

In both figures, it can be observed that the proposed \skipsideways (on the right) has a more stable training compared to \sideways. In particular, our \skipsideways is much more robust \wrt frame-rate. 
We hypothesise that through shortcut and direct connections, the network can build smoother representation, for instance, by interpolating between frames in the hidden space.
\noindent\textbf{Montage shots.}
We run experiments on
HMDB51, where we create training sequences by assembling a new video from two or more different clips, which we call \textit{montage shots}. Here, we investigate two scenarios. In the first scenario, we assemble 32 frames videos by concatenating 16 frames from one video and 16 frames from another video sampled from the same batch. Labels follow the same procedure. That is, the first 16 labels come from the first video, and the next 16 labels come from the second video. We also experiment with the concatenation of 3 video clips, each with 16 frames. Figure~\ref{fig:skip_sideways_montage_concat} shows the results. We can see a certain degree of robustness to cuts in a video clip. The more cuts, the worse the performance of the method. 

In the second scenario, we interleave frames from all videos. That is, we use the 1st frame from the video 1, then the 1st frame from the video 2, then the 1st frame from the video 3, then the 2nd frame from the video 1, then the 2nd frame from the video 2, then the 2nd frame from the video 3, and repeat that for a total of 48 frames, with 16 frames per clip. Note that the method, by design, is robust for interleaving 2 video clips as there is no interference between neurons operating on consecutive frames (e.g., $\bxt[1]$ and $\bxt[2]$ in Figure 1, in the main paper).  We did not manage to train the network with \skipsideways in that setting.

In both experiments, we use per-frame losses with per-frame labels. All the results above indicate that our \skipsideways training is robust if the input is reasonably smooth.

\subsection{Qualitative Results}

\begin{figure*}[t]
\begin{center}
\begin{tabular}{c@{\ }c@{\ }c@{\ }c@{\ }c@{\ }c@{\ }c}
\rotatebox{90}{\quad\quad\quad Input} &\includegraphics[width=0.17\linewidth]{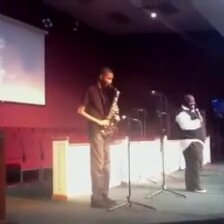} & 
\includegraphics[width=0.17\linewidth]{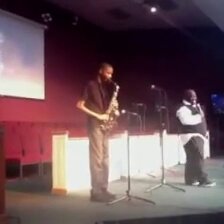} &
\includegraphics[width=0.17\linewidth]{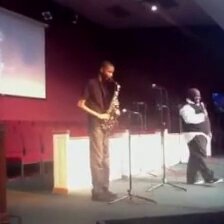} &
\includegraphics[width=0.17\linewidth]{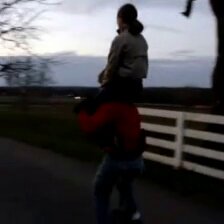} & 
\includegraphics[width=0.17\linewidth]{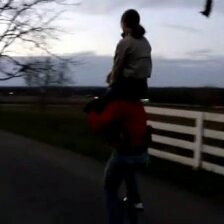} &
\includegraphics[width=0.17\linewidth]{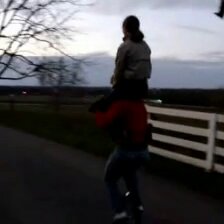}
\\
\rotatebox{90}{\quad Ground-Truth} &
\includegraphics[width=0.17\linewidth]{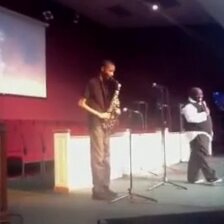} & 
\includegraphics[width=0.17\linewidth]{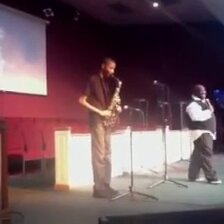} &
\includegraphics[width=0.17\linewidth]{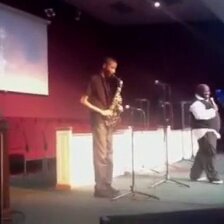} &
\includegraphics[width=0.17\linewidth]{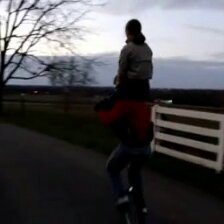} & 
\includegraphics[width=0.17\linewidth]{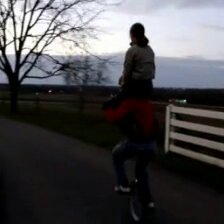} &
\includegraphics[width=0.17\linewidth]{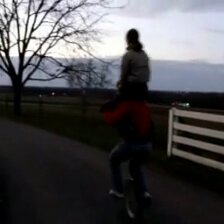} 
\\
\rotatebox{90}{\quad\quad \sideways} &
\includegraphics[width=0.17\linewidth]{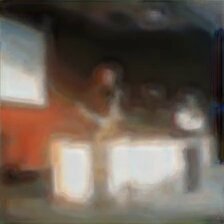} & 
\includegraphics[width=0.17\linewidth]{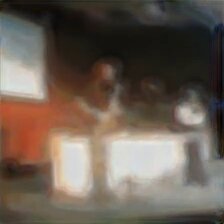} &
\includegraphics[width=0.17\linewidth]{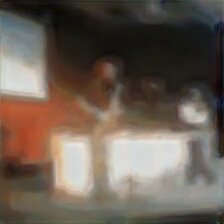} &
\includegraphics[width=0.17\linewidth]{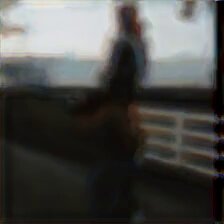} & 
\includegraphics[width=0.17\linewidth]{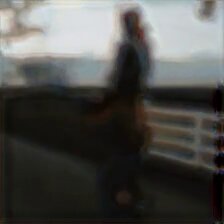} &
\includegraphics[width=0.17\linewidth]{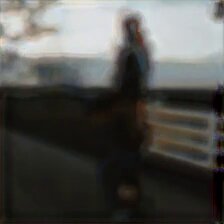} 
\\
\rotatebox{90}{\skipsideways (Ours)} &
\includegraphics[width=0.17\linewidth]{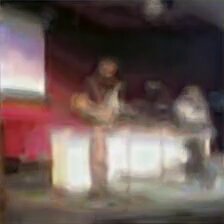} & 
\includegraphics[width=0.17\linewidth]{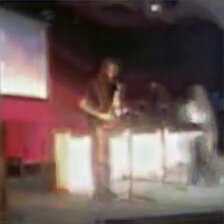} &
\includegraphics[width=0.17\linewidth]{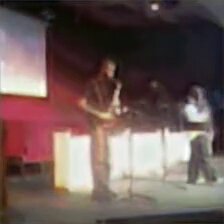} &
\includegraphics[width=0.17\linewidth]{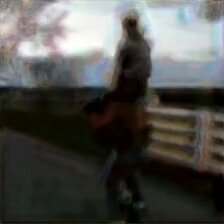} & 
\includegraphics[width=0.17\linewidth]{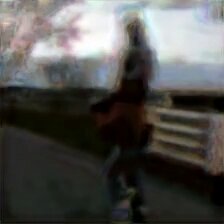} &
\includegraphics[width=0.17\linewidth]{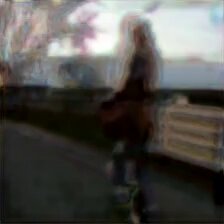}
\end{tabular}
\end{center}
\caption{Qualitative results for two video sequences from the validation fold of Kinetics600, where the models need to generate the frame at step $t+8$ into the future given frames up to $t$. From top to bottom: input frames, ground-truth output frames, predictions of \uniform  trained with \sideways, predictions of \uniform trained with \skipsideways.
We can observe that the predictions of the model trained with \sideways are blurry and semantically similar to inputs, indicating that the model lags behind. On the contrary, the predictions of the model trained with \skipsideways are sharper and semantically closer to the ground-truth output frames. We use resolution $224\times 224$.
}
\label{fig:future_forecasting_1}
\end{figure*}

We provide further qualitative results for the future frame prediction task. Here, we use higher frame resolution $224\times224$. Note that the same resolution is maintained over all the layers of our neural networks, from the input input towards the output.
We compare \sideways with \skipsideways to show the difference between both when the information is aggregated temporally. As noted in the main paper, using \uniform models is expensive for BP. This shows that we can possibly use computationally or memory-wise more expensive models in video modelling. We show the results in Figure~\ref{fig:future_forecasting_1}.

{\small\noindent\textbf{Acknowledgements.}
We thank Carl Doersch, Tom Hennigan, Jacob Menick, Simon Osindero, and Andrew Zisserman for their advice throughout the duration of the project and the anonymous CVPR reviewers for their feedback on the submission.
}
{\small
\bibliographystyle{ieee_fullname}
\bibliography{egbib}
}

\end{document}